\documentclass[10pt,journal,compsoc]{IEEEtran}

\usepackage{graphicx}
\usepackage{amsmath,amssymb} 
\usepackage{color}
\usepackage{verbatim}
\usepackage[leftcaption]{sidecap}
\usepackage{array}
\usepackage{times}
\usepackage{epsfig}
\usepackage{amsmath}
\usepackage{amssymb}
\usepackage{enumitem}
\usepackage{algorithm}
\usepackage{algorithmic}
\usepackage{bbding}
\usepackage{dsfont}
\usepackage{subcaption}
\usepackage{float}
\usepackage{tabularx}
\usepackage[numbers]{natbib}
\usepackage[pagebackref=true,breaklinks=true,letterpaper=true,colorlinks,bookmarks=false]{hyperref}
\usepackage{mathtools}
\usepackage{breqn}
\usepackage{multirow}

\newcommand{\modelname}{AOT-GAN}
\newcommand{\softdis}{SM-PatchGAN}
\newcommand{\harddis}{HM-PatchGAN}
\newcommand{\blockname}{AOT}
\newcommand{\etal}{et~al.} 
\newcommand{\etc}{etc.} 
\newcommand{\ie}{i.e.} 
\newcommand{\eg}{e.g.} 

\begin{document}
\title{
	Aggregated Contextual Transformations for High-Resolution Image Inpainting
	}
\author{Yanhong~Zeng,~\IEEEmembership{Student Member,~IEEE,}
	Jianlong~Fu,~\IEEEmembership{Member,~IEEE,}
	Hongyang~Chao,~\IEEEmembership{Member,~IEEE}
	and~Baining~Guo,~\IEEEmembership{Fellow,~IEEE}
\IEEEcompsocitemizethanks{
	\IEEEcompsocthanksitem This work has been submitted to the IEEE for possible publication. Copyright may be transferred without notice, after which this version may no longer be accessible.
	\IEEEcompsocthanksitem \copyright 2021 IEEE.  Personal use of this material is permitted.  Permission from IEEE must be obtained for all other uses, in any current or future media, including reprinting/republishing this material for advertising or promotional purposes, creating new collective works, for resale or redistribution to servers or lists, or reuse of any copyrighted component of this work in other works.

	\IEEEcompsocthanksitem This work is conducted when Yanhong~Zeng was a research intern in Microsoft Research. 
	\IEEEcompsocthanksitem Y. Zeng and H. Chao are with the School of Computer, Sun Yat-sen University, and also with Key Laboratory of Machine Intelligence and Advanced Computing, Ministry of Education, China. Email: zengyh7@mail2.sysu.edu.cn, isschhy@mail.sysu.edu.cn 
	\IEEEcompsocthanksitem J. Fu and B. Guo are with Microsoft Research. \protect\\
	Email: jianf@microsoft.com, bainguo@microsoft.com
	}
}


\IEEEtitleabstractindextext{%
\begin{abstract}
Image inpainting that completes large free-form missing regions in images is a promising yet challenging task. State-of-the-art approaches have achieved significant progress by taking advantage of generative adversarial networks (GAN). However, these approaches can suffer from generating distorted structures and blurry textures in high-resolution images (\eg, $512\times512$). The challenges mainly drive from (1) image content reasoning from distant contexts, and (2) fine-grained texture synthesis for a large missing region. To overcome these two challenges, we propose an enhanced GAN-based model, named \textbf{A}ggregated C\textbf{O}ntextual-\textbf{T}ransformation GAN (\textbf{\modelname}), for high-resolution image inpainting. 
Specifically, to enhance context reasoning, we construct the generator of \modelname~by stacking multiple layers of a proposed \blockname~block. The \blockname~blocks aggregate contextual transformations from various receptive fields, allowing to capture both informative distant image contexts and rich patterns of interest for context reasoning. 
For improving texture synthesis, we enhance the discriminator of \modelname~by training it with a tailored mask-prediction task. Such a training objective forces the discriminator to distinguish the detailed appearances of real and synthesized patches, and in turn facilitates the generator to synthesize clear textures.
Extensive comparisons on Places2, the most challenging benchmark with 1.8 million high-resolution images of 365 complex scenes, show that our model outperforms the state-of-the-art by a significant margin in terms of FID with \textbf{38.60\% relative improvement}. 
A user study including more than \textbf{30 subjects} further validates the superiority of \modelname.
We further evaluate the proposed \modelname~in practical applications, \eg, logo removal, face editing, and object removal. Results show that our model achieves promising completions in the real-world.
We release code and models in \url{https://github.com/researchmm/AOT-GAN-for-Inpainting}. 
\end{abstract}
}

\maketitle
\IEEEdisplaynontitleabstractindextext
\IEEEpeerreviewmaketitle

\IEEEraisesectionheading{
	\section{Introduction}
	\label{sec:introduction}}

\IEEEPARstart{I}{mage} inpainting aims at filling missing regions in images with plausible contents \cite{Bertalmio2000image}. An effective image inpainting algorithm plays an important role in numerous applications, such as object removal, photograph restoration and diminished reality \cite{barnes2009patchmatch,criminisi2004region,wexler2007space,dimini2015,pixmix2014}. Despite of the great benefits of this technology, image inpainting meets grand challenges in simultaneously recovering reasonable contents and clear textures for large free-form missing regions in high-resolution images.

\begin{figure}[!t]
\centering		
\includegraphics[width=\linewidth]{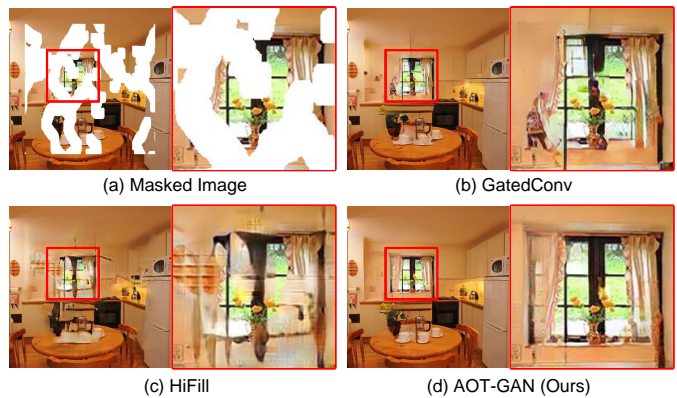} 
\caption{
	An illustration of image inpainting. 
	Given an image  (a) (resolution: $512\times 512$) with large irregular holes, our model is able to reconstruct more plausible structures and clearer textures of the window in (d) compared with GatedConv \cite{yu2019free} (b) and HiFill \cite{yi2020contextual} (c). We present enlarged patches by red boxes next to the images.}
\label{fig:teaser}
\end{figure}

Early works attempt to solve the problem by either diffusion-based or patch-based algorithms \cite{Bertalmio2000image,barnes2009patchmatch,criminisi2004region,bertalmio2003simultaneous,mvv2018,patch2017}. A diffusion-based algorithm propagates contextual information from boundaries into holes in the isophotes direction \cite{Bertalmio2000image,bertalmio2003simultaneous}. A patch-based algorithm synthesizes missing contents by copying and pasting similar patches from uncorrupted image contexts or external database \cite{barnes2009patchmatch,criminisi2004region,wexler2007space,sun2005image}. These approaches can work well in completing narrow holes in stationary backgrounds. However, they often fail to hallucinate plausible contents in semantic inpainting due to the lack of powerful reasoning for missing contents and textures in complex scenes \cite{yu2018generative,pathak2016context}.  

Significant progress has been made by the emergence of deep feature learning \cite{he2016deep} and adversarial training \cite{goodfellow2014generative} for image inpainting \cite{yu2018generative,pathak2016context,nazeri2019edgeconnect,zeng2020learning}. These deep image inpainting models are typically built upon generative adversarial networks (GAN) \cite{goodfellow2014generative}. 
Through a joint optimization by reconstruction losses and adversarial losses \cite{yu2019free,yu2018generative}, existing deep models have shown promising results in semantic inpainting. 
However, directly applying these deep models in higher-resolution images (e.g., $512\times512$) often results in distorted structures and blurry textures \cite{yang2017high,wang2018image}. 
Such a problem hinders these algorithms from practical applications, where users' photos are often larger than $256\times256$.
The difficulties mainly drive from simultaneously inferring reasonable contents from distant image contexts and generating fine-grained image presentation for each pixel over a large region. We discuss these two challenges respectively as below. 

\textbf{Challenge 1: Context reasoning.} 
To infer reasonable contents for missing regions, deep image inpainting models leverage features from distant image contexts. 
For example, contextual attention modules are proposed to model relations between hole regions and image contexts by patch-wise matching \cite{yu2018generative,zeng2019learning}. However, the patch-wise matching often results in distorted structures due to the issue of repeating patterns \cite{liu2019coherent}. 
The other line of works stack serialized dilated convolution layers to capture distant contexts. However, as widely discussed in previous works \cite{yu2018generative,wang2018understanding}, serialized dilated convolutions tend to encode features of predefined gridding patterns rather than capturing rich patterns of interest for context reasoning. 
As shown in Fig. \ref{fig:teaser}, the state-of-the-art approaches, GatedConv \cite{yu2019free} (b) and HiFill \cite{yi2020contextual} (c), fail in completing the structures of the window in results.

\textbf{Challenge 2: Fine-grained texture synthesis.}
To synthesize fine-grained textures, state-of-the-art models utilize adversarial training by designing a game-theoretical minimax game between a generator and a discriminator. 
The majority of them inherit the discriminator from PatchGAN \cite{isola2017image} due to its great success in image translation \cite{nazeri2019edgeconnect,isola2017image,xiong2019foreground}. 
The discriminator of PatchGAN aims to distinguish the patches of a real image from those of an inpainted result.   
However, they ignore the fact that the patches outside missing regions are shared by both real and inpainted images, and blindly push the discriminator to distinguish these identical patches. Such a problematic learning objective weakens the discriminator, and in turn hinders the generator from synthesizing realistic fine-grained textures. As shown in Fig. \ref{fig:teaser}, GatedConv \cite{yu2019free} (b) and HiFill \cite{yi2020contextual} (c) tend to generate unrealistic textures in the missing regions. 

To overcome the above two challenges, we propose a novel GAN-based model for high-resolution image inpainting, named \textbf{A}ggregated C\textbf{O}ntextual-\textbf{T}ransformation GAN (\modelname). It consists of (1) a generator that exploits aggregated contextual transformations of distant image contexts for enhancing context reasoning, and (2) a discriminator trained by a tailored mask-prediction task to facilitate the synthesis of fine-grained textures. 

To enhance context reasoning for large missing regions, we build the generator of \modelname~by repeating a carefully-designed building block, namely \blockname~block.
Specifically, the \blockname~blocks adopt the \textit{split-transform-merge} strategy. First, an \blockname~block splits the kernel of a standard convolution into multiple sub-kernels. Second, it applies each sub-kernel on the incoming features with different dilation rates. Finally, the \blockname~block aggregates different transformations from all the sub-kernels as output (Fig. \ref{fig:block}-(b)).
Through these three steps, informative distant image contexts are leveraged by using various dilation rates, while rich patterns of interest are captured by aggregating multiple transformation results, leading to better context reasoning for missing regions. 
To the best of our knowledge, \blockname~block is the first highly-modularized building block tailored for high-resolution image inpainting.

To facilitate the synthesis of fine-grained textures, we design a novel mask-prediction task to train the discriminator. 
Most existing approaches follow PatchGAN \cite{patch2017} to force the discriminator to predict all patches in inpainted images as fake, while ignoring the fact that those outside missing regions indeed come from real images. Therefore, these approaches may struggle to generate realistic fine-grained textures. 
To overcome the above limitation, we instead train the discriminator to predict a downsampled patch-level inpainting mask. In other words, for an inpainted image, the discriminator is expected to segment the synthesized patches from real ones. Such a learning objective leads to a stronger discriminator and in turn improve the generator to synthesize fine-grained realistic textures.  

We summarize our contributions as follows:
\begin{itemize}[nosep]
	\item  We propose to learn aggregated contextual transformations for high-resolution image inpainting, which allows capturing both informative distant contexts and rich patterns of interest for context reasoning.

	\item We design a novel mask-prediction task to train the discriminator tailored for image inpainting. Such a design forces the discriminator to distinguish the detailed appearances of real and synthesized patches, which in turn facilitates the generator to synthesize fine-grained textures.
	
	\item We conduct extensive evaluations on the most challenging benchmark, Places2 \cite{zhou2018places}. Results show that our model outperforms the state-of-the-art by a significant margin in terms of FID with 38.60\% relative improvement.
\end{itemize}

The remainder of the paper is organized as follows. We review related works in Section \ref{sec:related}, and introduce our approach in Section \ref{sec:approach}. Extensive experiments are presented in Section \ref{sec:experiment}, followed by the conclusions in Section \ref{sec:conclusion}.

\section{Related work}
\label{sec:related}

Image inpainting aims at filling missing regions in images with plausible contents \cite{Bertalmio2000image}. Due to its significant practical values for image editing applications (\eg, object removal and photograph restoration), image inpainting has become an active research topic for decades \cite{barnes2009patchmatch,criminisi2004region,yu2018generative,pathak2016context,nazeri2019edgeconnect,zeng2019learning}. Existing approaches can fall into two categories, \ie, non-learning based or learning-based methods. In this section, we discuss these approaches as below. 

\subsection{Non-learning based Image Inpainting}
Prior to the prevalence of deep learning, approaches for image inpainting are typically divided into two categories, diffusion-based and patch-based algorithms \cite{Bertalmio2000image,barnes2009patchmatch,criminisi2004region,bertalmio2003simultaneous,sun2005image}. We introduce both of them in this section. 

A diffusion-based algorithm propagates contextual pixels from boundaries to the holes along the isophotes direction \cite{Bertalmio2000image,bertalmio2003simultaneous,ballester2001variational, bertalmio2001navier}. Specifically, many boundary conditions are imposed by the use of Partial Differential Equations during the pixel propagation. Through leveraging the prior knowledge of these boundary conditions, diffusion-based approaches have shown promising results in filling in narrow sketches. However, these approaches typically introduce diffusion-related blurs thus can fail in completing large missing regions \cite{Bertalmio2000image,bertalmio2003simultaneous,ballester2001variational, bertalmio2001navier}. 

\begin{figure*}[!t]
  \centering
  \includegraphics[width=\linewidth]{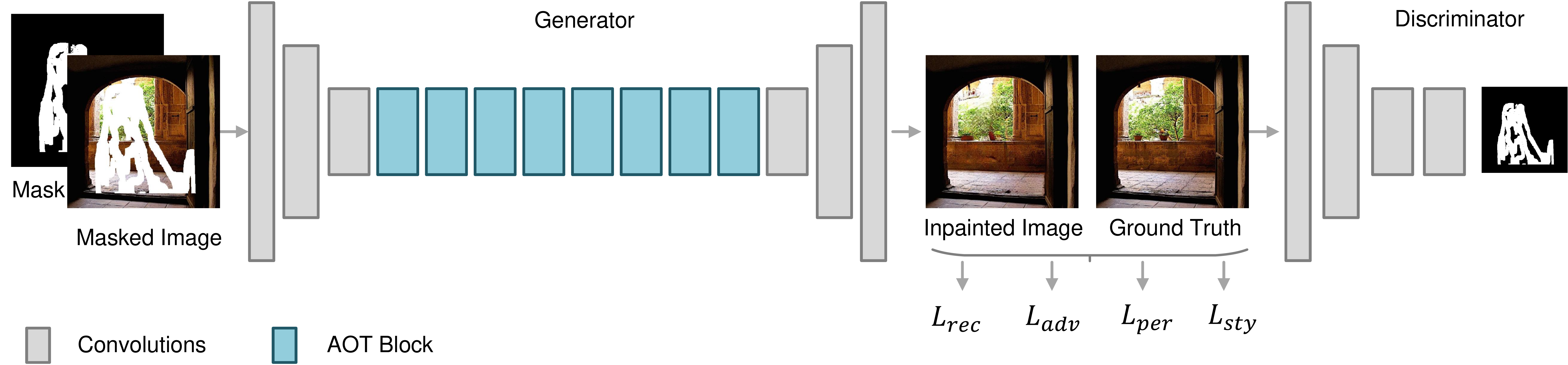}
	\caption{The overview of the proposed \textbf{A}ggregated C\textbf{O}ntextual-\textbf{T}ransformation GAN (\modelname). \modelname~is built upon a generative adversarial network (GAN), which consists of a generator and a discriminator. 
	Specifically, the generator is highly-modularized by stacking multiple-layers of a carefully-designed block, \ie, \blockname~block, for enhancing context reasoning. 
	The discriminator is trained by a tailored mask-prediction task, which aims at predicting downsampled patch-level inpainting masks. 
	Our model is jointly optimized by a reconstruction loss, an adversarial loss \cite{goodfellow2014generative}, a perceptual loss \cite{johnson2016perceptual} and a style loss \cite{gatys2016image}.
	Details can be found in Section \ref{sec:approach}.}
	\label{fig:framework}
\end{figure*}

Inspired by patch-based methods for texture synthesis \cite{liang2001real,efros1999texture}, more and more patch-based filling approaches are proposed for image inpainting \cite{criminisi2004region,efros2001image}. A patch-based approach typically synthesizes missing contents by copying and pasting similar patches from known image contexts or external database. To improve the patch-based algorithm, many efforts have been made to find the optimal patch size, patch offset, filling orders, and matching algorithms \cite{barnes2009patchmatch,criminisi2004region,wexler2007space,sun2005image}. 
For example, Criminisi \etal~employ an exemplar-based texture synthesis technique and use calculated confidence values together with image isophotes for determining the filling order of the target inpainting region \cite{criminisi2004region}. Sun \etal~propose a curve-based interactive approach to complete important structures before remaining unknown regions \cite{sun2005image}. Barnes \etal~propose a new randomized algorithm, namely PatchMatch, for quickly finding approximate nearest neighbor matches between image patches \cite{barnes2009patchmatch}. 
These approaches can work well especially in the stationary background inpainting with repeating patterns. However, these approaches can fall short in completing large missing regions of complex scenes for semantic inpainting. This is because patch-based approaches heavily rely on patch-wise matching by low-level features (\eg, isophotes). Such a technology is unable to synthesize contents for which similar patches do not exist in known image contexts. 

\subsection{Learning-based Image Inpainting}
Significant progress has been made by the emergence of deep feature learning \cite{he2016deep} and adversarial training \cite{goodfellow2014generative} for image inpainting \cite{yu2018generative,pathak2016context,nazeri2019edgeconnect,iizuka2017globally}. Compared with non-learning based approaches, deep image inpainting models are able to synthesize more plausible contents for complex scenes. 

To infer missing contents for a large missing region, deep image inpainting models take advantage of features from distant image contexts. 
Context Encoder \cite{pathak2016context}, a seminal deep inpainting model, adopts a fully-connected layer in a compact latent feature space for global image contexts encoding. Early works that follow the architecture of Context Encoder have shown promising results on images of human faces, street views, \etc~\cite{yang2017high,li2017generative,yeh2017semantic}. 
However, these models can only deal with low-resolution images (\eg, $128\times 128$) and missing regions of fixed shapes (\eg, center squares) due to the usage of fully-connected layers. To solve this problem, Iizuka \etal~propose to build models upon fully convolutional networks (FCN) \cite{iizuka2017globally}. 
To capture distant contexts based on FCN, a contextual attention module is proposed to find patches of interest from contexts by patch-wise matching \cite{yu2018generative, liu2019coherent, yan2018shift}. Zeng \etal~futher adopt a non-local module named attention transfer network to fill missing regions in feature pyramids \cite{zeng2019learning}. Despite the effective long-range relation modeling, non-local modules tend to repeat patterns in missing regions and distort structures \cite{liu2019coherent,yan2018shift,xie2019image,song2018contextual}. To avoid the issue of repeating patterns, the other line of works stack serialized dilated convolution layers to capture distant contexts \cite{yu2018generative,wang2018image,sagong2019pepsi}. Dilated convolutions use kernels that are spread out, allowing to compute each output pixel with a much larger receptive field than standard convolutions \cite{yu2016multi,chen2017deeplab}. By stacking the dilated convolutions, the model can effectively ``see'' a larger area of the input image for context reasoning. However, as widely acknowledged, serialized dilated convolutions tend to encode features of predefined gridding patterns rather than patterns of interest for context reasoning \cite{yu2018generative,wang2018understanding,wang2018smoothed}.

Inspired by recent works of style transfer \cite{johnson2016perceptual,gatys2016image}, increasing numbers of deep inpainting models exploit a style loss \cite{gatys2016image} and a perceptual loss \cite{johnson2016perceptual} for the synthesis of fine-grained textures. Specifically, the joint optimization of a style loss and a perceptual loss aims at minimizing the perceptual distance between the deep features of inpainted results and the original images. Although promising results have been shown, it remains challenging for these models to generate clear textures \cite{liu2018image}. 

Significant progress has been made by adopting the framework of generative adversarial networks (GAN) for semantic inpainting \cite{goodfellow2014generative}. A GAN-based model typically consists of a generator network and a discriminator network. The discriminator is trained to distinguish inpainted images from real ones, while the generator is optimized to synthesize realistic images to fool the discriminator. Through the game-theoretical minmax game between the generator and the discriminator, GAN-based inpainting models are able to generate clearer textures. 
To further improve the discriminator network, Iizuka \etal~propose a join training by a global and a local discriminator \cite{iizuka2017globally}. The global discriminator looks at the entire image and the local discriminator focuses on the local consistency. However, the local discriminator can only deal with missing regions of fixed shapes due to the usage of fully-connected layers in its network. 
To solve this problem, Yu \etal~inherit the discriminator from PatchGAN \cite{isola2017image} due to its great success in image translation \cite{yu2019free,nazeri2019edgeconnect,yu2018generative,zeng2019learning,xiong2019foreground,ren2019structureflow}. The discriminator of PatchGAN aims to distinguish patches of real images from those of inpainted results. Moreover, they apply spectral normalization to each layer of the discriminator to stabilize the training of GAN \cite{miyato2018spectral}. 
However, PatchGAN-based models usually ignore the fact that patches outside missing regions are shared by both real and inpainted images, and blindly pushing the discriminator to distinguish these identical patches can weaken the discriminator.


\section{Approach}
\label{sec:approach}

Inferring reasonable contents and generating clear textures for a large free-form missing region are vital yet challenging to high-resolution image inpainting. 
In this section, we present the details of the proposed \textbf{A}ggregated C\textbf{O}ntextual-\textbf{T}ransformation GAN (\modelname), which enhances the context reasoning and texture synthesis for high-resolution image inpainting. 
As shown in Fig. \ref{fig:framework}, AOT-GAN is built upon a generative adversarial network (GAN), which consists of a generator network and a discriminator network. Specifically, the generator is constructed by stacking multiple layers of a carefully-designed building block, \ie~\blockname~block. The \blockname~block is able to capture both informative distant contexts and rich patterns of interest for enhancing context reasoning. The discriminator is improved by a tailored mask-prediction task, which can in turn facilitates the generator to synthesize clear textures. \modelname~is trained by a joint optimization of a reconstruction loss, an adversarial loss, a perceptual loss and a style loss. Through such a joint optimization, \modelname~is able to synthesize reasonable contents and clear textures for missing regions of high-resolution images. 

In this section, we introduce the details of \blockname~block in Section \ref{sec:block} and the design of mask-prediction task for the discriminator in Section \ref{sec:dis}. The overall optimization objectives are introduced in Section \ref{sec:loss}, followed by the implementation details in Section \ref{sec:imp}.

\subsection{Aggregated Contextual Transformations}
\label{sec:block}
In this section, we present a simple and highly-modularized generator network, which is enhanced for context reasoning. In particularly, context reasoning for a large free-form missing region is one of the grand challenges to high-resolution image inpainting \cite{nazeri2019edgeconnect,ren2019structureflow}. For one thing, to ensure a coherent structure with surrounding image contexts, deep image inpainting models need to propagate information from distant contexts to missing regions. For another, as objects in complex scenes have various scales and angles of view, capturing as rich as possible patterns of interest is important for context reasoning. The \blockname~block is designed to capture both informative distant image contexts and rich patterns of interest for context reasoning in high-resolution image inpainting. 

The overview of the generator network of \modelname~is depicted in Fig. \ref{fig:framework}. Specifically, the generator consists of an encoder, a stack of the building blocks, \ie~\blockname~blocks, and a decoder. The generator takes as inputs a masked image and a binary mask indicating missing regions, and it outputs a completed image. First, we stack several layers of standard convolutions as the encoder for feature encoding from the input. Second, features from the encoder are processed by the stack of \blockname~blocks. Through the backbone built by \blockname~blocks, the generator is able to capture informative distant image contexts as well as rich patterns of interest for filling missing regions. Finally, we use multiple layers of transposed convolutions to decode features from \blockname~blocks to a completed image as the final result. We discuss more details of the design of \blockname~blocks as below. 

\begin{figure}[!t]
	\centering		
	\includegraphics[width=\linewidth]{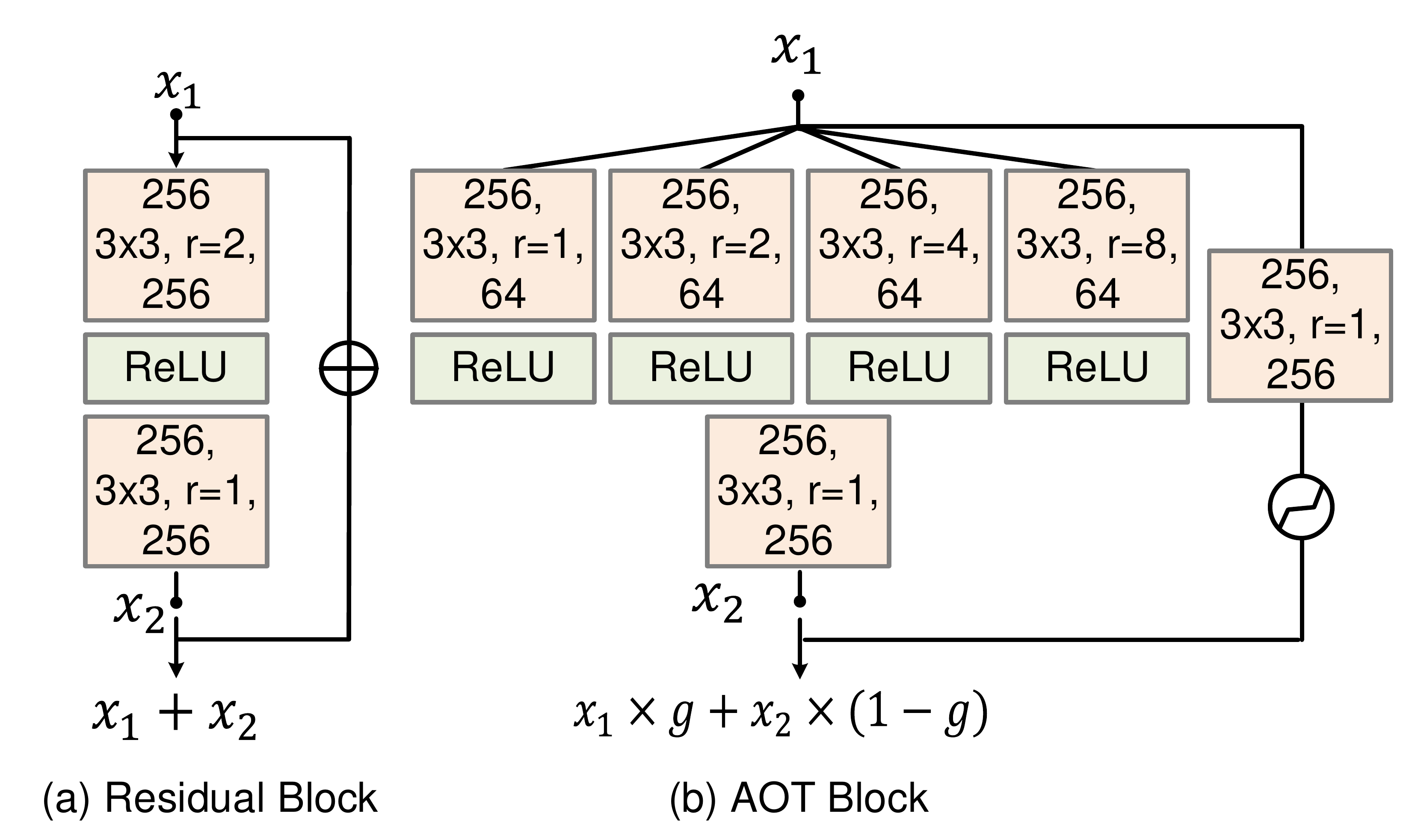}
	  \caption{An illustration of the Residual block (a) used in the state-of-the-art deep inpainting models \cite{nazeri2019edgeconnect,xie2019image,ren2019structureflow} and the proposed \blockname~block (b). The numbers inside orange blocks denote (\#input channels, filter sizes, dilation rates, \#output channels). }
	  \label{fig:block}
  \end{figure}

\blockname~blocks adopt the \textit{split-transformation-merge} strategy by three steps \cite{xie2017aggregated}. \textit{(i) Splitting}: as shown in Fig. \ref{fig:block}-(b), an \blockname~block splits the kernel of a standard convolution into multiple sub-kernels, each of which has fewer output channels. For example, splitting a kernel with 256 output channels into four sub-kernels makes each sub-kernel has 64 output channels. \textit{(ii) Transforming}: each sub-kernel performs a different transformation of the input feature $x_1$ by using a different dilation rate. Specifically, a dilated kernel is spread out by introducing zeros between consecutive positions, which has achieved a great success in the field of semantic segmentation \cite{yu2016multi}. Using a larger dilation rate enables the sub-kernel to ``see'' a larger area of the input image, while a sub-kernel that uses a small dilation rate focuses on the local patterns of a smaller receptive field. 
\textit{(iii) Aggregating}: the contextual transformations from different receptive fields are finally aggregated by a concatenation followed by a standard convolution for feature fusion. Such a design allows the \blockname~block to predict each output pixel through different views. Through the above three steps, an \blockname~block is able to aggregate multiple contextual transformations for enhancing context reasoning. 

Furthermore, the stack of \blockname~blocks largely enriches the combinations of different pathways in the generator network, allowing the generator to capture as many as possible patterns of interest for context reasoning. It is worth noting that, compared with a standard residual block used in the state-of-the-art deep inpainting models (Fig. \ref{fig:block}-(a)), \blockname~blocks do not introduce extra model parameters and computation costs in these three steps.

Inspired by the great success of ResNet \cite{he2016deep}, deep inpainting models typically incorporate an identical residual connection in their building blocks to ease the training of networks. As shown in Fig. \ref{fig:block}-(a), an identical residual connection aggregates the input feature $x_1$ and the learned residual feature $x_2$ by element-wise summing in a spatially-invariant way. However, they ignore the discrepancies between the values of input pixels inside and outside missing regions, resulting in the problem of color discrepancy in inpainted images \cite{yu2019free,liu2018image}. To 
alleviate this problem, we propose to use a novel gated residual connection in the building block. As shown in Fig. \ref{fig:block}-(b), a gated residual connection first calculates the spatially-variant gate value $g$ from $x_1$ by a standard convolution and a sigmoid operation, and then the \blockname~block aggregates the input feature $x_1$ and the learned residual feature $x_2$ by a weighted sum with $g$, which is denoted as:
\begin{equation}
		x_3 = x_1 \times g + x_2 \times (1-g).
\end{equation} 
Such a spatially-variant feature aggregation encourages updating features inside missing regions while retaining the known features outside missing regions. We conduct both quantitative and qualitative experiments to show the improvements by the gated residual connection in Section \ref{sec:ab-gate}. 

\textit{Relation to residual blocks \cite{he2016deep}.}
To enlarge the receptive fields, the state-of-the-art inpainting models also improve the residual block by using dilated convolutions as their building blocks (Fig. \ref{fig:block}-(a)) \cite{nazeri2019edgeconnect,xie2019image,ren2019structureflow}. Specifically, they use the same dilation rate for the kernel inside one building block. Although distant image contexts can be leveraged by the stack of such an improved residual block, it often suffers from encoding features of predefined gridding patterns instead of patterns of interest for context reasoning \cite{yu2018generative,wang2018understanding,wang2018smoothed}. On the contrast, the proposed \blockname~block is able to capture rich patterns of interest by learning aggregated contextual transformations with various dilation rates. 

\textit{Relation to Atrous Spatial Pyramid Pool (ASPP) \cite{yu2016multi}.}
ASPP is a multi-branch module that shares a similar topology of the proposed \blockname~block. We highlight the difference that, ASPP is proposed for the task of semantic segmentation and it is used at the end of networks for high-level recognition, while \blockname~block is a basic building block repeating in inpainting models for low-level tasks. Besides, the \blockname~block incorporates a carefully-designed gated residual connection that cares about the problem of color discrepancy for image inpainting.

\subsection{Soft Mask-Guided PatchGAN (\softdis)}
\label{sec:dis}
In this section, we present a novel mask-prediction task for enhancing the training of the discriminator of \modelname, which can in turn facilitate the generator to synthesize clear textures for high-resolution image inpainting. 

Synthesizing clear fine-grained textures for a large missing region is another challenge for high-resolution image inpainting. This is because there are various possible results for filling a missing region, whereas deep inpainting models are trained to reconstruct the original images based on reconstruction losses (\eg, $L_1$ loss) \cite{yu2018generative}. With such an optimization objective, deep inpainting models tend to generate an average of all the possible solutions, resulting in blurry textures. To overcome this challenge, we build our model upon a generative adversarial network (GAN) \cite{goodfellow2014generative}. Through the joint optimization of reconstruction losses and an adversarial loss, our model, \modelname, is able to generate clearer results that fall near the manifold of natural images. 

In practice, we denote a real image as $x$, a corresponding binary inpainting mask as $m$ (with value 0 for known pixels and 1 for missing regions), $\odot$ as pixel-wise multiplication, $G$ as the generator, and thus the inpainted results can be denoted as:
\begin{equation}
	z = x\odot (1-m) + G(x\odot(1-m), m) \odot m.
\end{equation} 
We inherit the network architecture of the discriminator from PatchGAN \cite{isola2017image} due to its great success in the field of image translation. As shown in Fig. \ref{fig:framework}, the discriminator network consists of several layers of standard convolutions, each of which reduces the spatial size of feature maps by two. The discriminator takes as input an image from inpainted results or real data, and outputs a prediction map. In particular, each pixel of the prediction map indicates the prediction for a $N\times N$ patch in the input image as real or fake.  

\begin{figure}[!t]
	\centering
	\includegraphics[width=\linewidth]{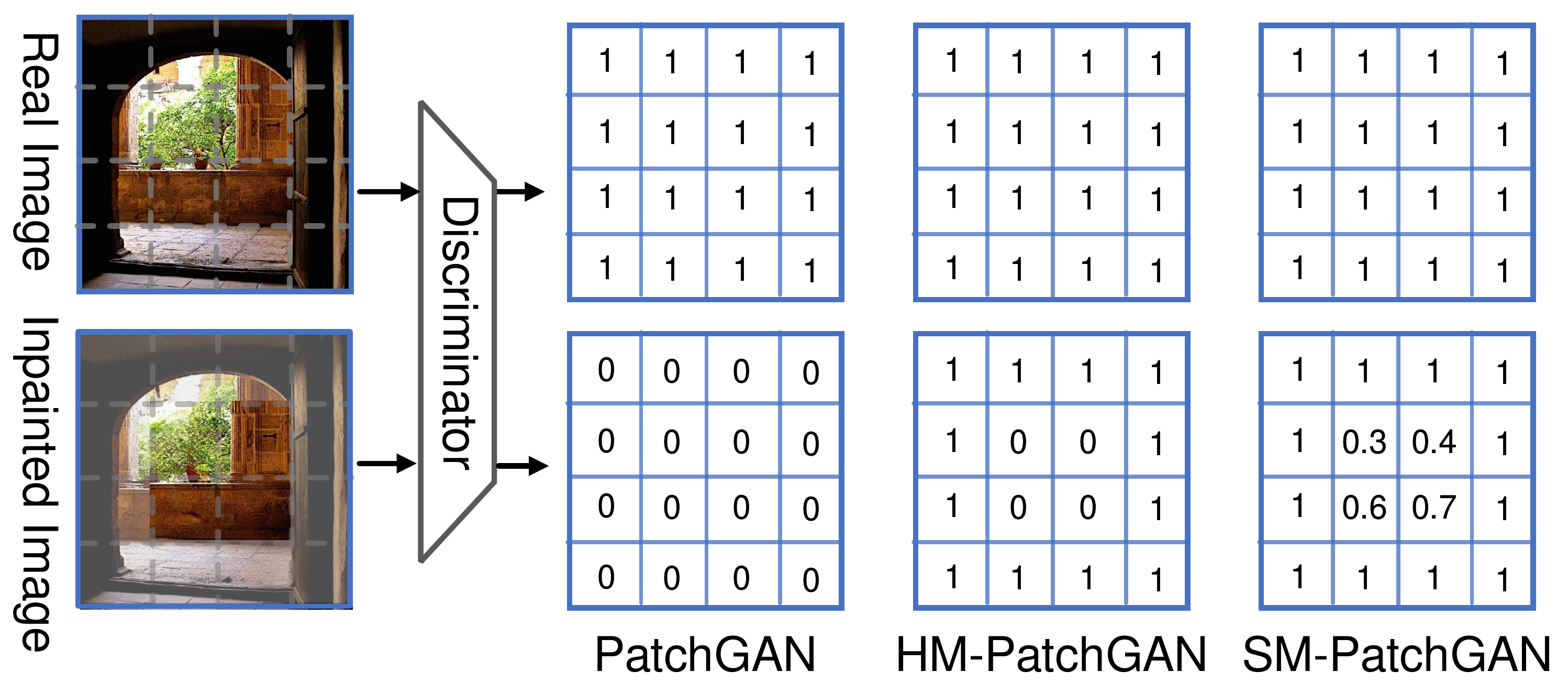}
	  \caption{An illustration of different tasks for the training of the discriminator. PatchGAN aims at distinguishing patches of inpainted images from those of real images, while HM-PatchGAN and SM-PatchGAN aim to segment synthesized patches of missing regions from real ones of contexts according to inpainting masks. 
	  }
	  \label{fig:dis}
  \end{figure}

To facilitate the synthesis of fine-grained textures, we design a mask-prediction task for the training of the discriminator, which we term a Soft Mask-guided PatchGAN (\softdis). Specifically, we down sample the inpainting masks as the ground truths of the mask-prediction task. To simulate the pixel propagation around the boundaries of missing regions, as shown in Fig. \ref{fig:dis}, we use a soft patch-level mask for training. The soft mask is obtained by Gaussian filtering. We denote the adversarial loss for the discriminator as:
\begin{dmath}
	L_{adv}^D = \mathbb{E}_{z \sim p_z} \left[(D(z) - \sigma(1-m))^2\right] + \\ \mathbb{E}_{x \sim p_{data}}\left[(D(x)-1)^2 \right],
\end{dmath}
where $\sigma$ denotes the composition function of the down-sampling and the Gaussian filtering. Correspondingly, the adversarial loss for the generator is denoted as:
\begin{dmath}
	L_{adv}^G =\mathbb{E}_{z \sim p_z} \left[ (D(z) -1 )^2 \odot m \right],
	\label{eq:advg}
\end{dmath} 
where only predictions of the synthesized patches for missing regions are used to optimize the generator. Through such an optimization, the discriminator is encouraged to segment synthesized patches of missing regions from real ones of contexts outside missing regions. The enhanced discriminator can in turn facilitate the generator to synthesize more realistic textures. 

\textit{Relation to PatchGAN \cite{isola2017image}.}
As shown in Fig. \ref{fig:dis}, the PatchGAN discriminator used in previous inpainting models aims at distinguishing patches of inpainted images from those of real ones. In this way, they blindly push the discriminator to predict all patches in inpainted images as fake, while ignoring that patches outside missing regions indeed come from real images. In contrast, the proposed \softdis~aims at distinguishing synthesized patches of missing regions from real ones of contexts, which can lead to a stronger discriminator. 

\textit{Relation to HM-PatchGAN.} We consider another possible design of the mask-prediction task, which we term a Hard Mask-guided PatchGAN (\harddis). As shown in Fig. \ref{fig:dis}, the \harddis~enhances the PatchGAN discriminator by training by hard binary inpainting masks without Gaussian filtering. The \harddis~ignores the fact that a patch in an inpainted image around boundaries may contain both real and synthesized pixels. We speculate that such a design can weaken the training of the discriminator. To avoid the above issue, the proposed \softdis~instead uses soft labels by the processing of Gaussian filtering. We conduct extensive ablation study to show the superiority of \softdis~in Section \ref{sec:ab-dis}.

\subsection{Overall Optimization}
\label{sec:loss}
The principle of choosing optimization objectives for image inpainting is to ensure both per-pixel reconstruction accuracy and visual fidelity of inpainted images. To this end, we carefully select four optimization objectives, \ie, a $L_1$ loss, a style loss \cite{gatys2016image}, a perceptual loss \cite{johnson2016perceptual}, and an adversarial loss of the proposed \softdis, for \modelname~following the majority of existing deep inpainting models \cite{yu2018generative,nazeri2019edgeconnect,yan2018shift,liu2018image}. First, we include a $L_1$ loss to ensure the reconstruction accuracy on pixel-level: 
\begin{equation}
L_{rec} = \left \| x - G(x\odot(1-m), m) \right\|_1.
\end{equation}
Since the effectiveness of the perceptual loss \cite{johnson2016perceptual} and the style loss \cite{gatys2016image} for image inpainting have been widely verified \cite{nazeri2019edgeconnect,liu2018image}, we include them to improve the accuracy of perceptual reconstruction. Specifically, a perceptual loss aims at minimizing the $L_1$ distance between the activation maps of inpainted and real images:
\begin{equation}
	L_{per} = \sum_{i} \frac{\|\phi_i(x) - \phi_i(z)\|_1}{N_i} ,
\end{equation} 
where $\phi_i$ is the activation map from the $i$-th layers of a pre-trained network (e.g., VGG19 \cite{simonyan2014very}), $N_i$ is the number of elements in $\phi_i$. Similarly, the Style loss is defined as the $L_1$ distance between the Gram matrices of deep features of inpainted and real images: 
\begin{equation}
	L_{sty} = \mathbb{E}_i \left [ \|\phi_i(x)^T\phi_i(x) - \phi_i(z)^T\phi_i(z)\|_1 \right ]. 
\end{equation}
Finally, we include the adversarial loss of \softdis~described in Eq. \ref{eq:advg} to improve the visual fidelity of inpainted images. 
The whole \modelname~is trained by a joint optimization of these four objectives, and we conclude the overall optimization objective as below:
\begin{dmath}
L = \lambda_{adv}L_{adv}^G + \lambda_{rec}L_{rec} + \lambda_{per}L_{per} + \lambda_{sty}L_{sty}. 
\end{dmath}
For our experiments, we empirically choose $\lambda_{adv}=0.01$, $\lambda_{rec}=1$, $\lambda_{per} = 0.1$ and $\lambda_{sty}=250$ for training.

\subsection{Implementation Details} 
\label{sec:imp}
To build the generator network of the proposed \modelname, we stack three layers of convolutions as the encoder, which reduces the spatial size of images by four times in total. Correspondingly, we use three layers of transposed convolutions for the upsampling in the decoder. We stack eight layers of \blockname~blocks to build the backbone of the generator. For building the discriminator network, we stack four layers of convolutions with stride as two following $70 \times 70$ PatchGAN \cite{isola2017image}. Accordingly, for the processing of Gaussian filtering in \softdis, we set the kernel size of the Gaussian kernel as $70\times 70$ to simulate the pixel propagation by convolutions in the discriminator network. 
In particular, to avoid the issue of color shift caused by normalization layers, we remove all normalization layers in the generator network following previous works \cite{yu2019free,yu2018generative}. 

For the model training, eight images are randomly sampled and corresponding masks are randomly created as pairs in each mini-batch. We use a fixed learning rate at $1e-4$ for both the training of the discriminator and the generator. We use Adam optimizer with $\beta_1=0$ and $\beta_2=0.9$ for training. 
In practice, we use VGG19 \cite{simonyan2014very} pretrained on ImageNet \cite{deng2009imagenet} as our pre-trained network for calculating the style loss and the perceptual loss. Code and models are available in  \url{https://github.com/researchmm/AOT-GAN-for-Inpainting}.


\section{Experiments}
\label{sec:experiment}
In this section, we introduce the details of datasets, baselines and evaluation metrics for image inpainting in Section \ref{sec:data}-\ref{sec:metric}. We conduct both quantitative and the qualitative evaluations in Section \ref{sec:quant}  and Section \ref{sec:qual}, followed by a user study in Section \ref{sec:user}. Extensive ablation studies are introduced in Section \ref{sec:ab} to verify the effectiveness of each component of the \modelname. Finally, we show practical applications of our model in Section \ref{sec:application} and discuss limitations in Section \ref{sec:limit}.

\subsection{Datasets}
\label{sec:data}
We use free-form masks provided by Liu \etal~\cite{liu2018image} for both training and testing following common settings \cite{yu2019free,liu2019coherent,liu2018image,yi2020contextual}. Because the free-form masks are more challenging, closer to real-world applications and also widely adopted by a majority of inpainting approaches \cite{nazeri2019edgeconnect,liu2018image}. All images are resized to $512\times 512$ for both training and evaluation following a standard high-resolution setting \cite{yang2017high,song2018contextual}. 
We introduce three datasets used in this paper with their brief introductions as below:
\begin{itemize}
	\item \textbf{Places2} \cite{zhou2018places} contains over 1.8 million images from 365 kinds of scenes. It is one of the most challenging datasets for image inpainting due to its complex scenes. We use the splits of training/testing (\ie, 1.8 million/36,500) following the settings used by most inpainting models \cite{yu2018generative,nazeri2019edgeconnect,zeng2019learning}. 
	
	\item \textbf{CELEBA-HQ} \cite{karras2018pggan} is a high-quality dataset of human faces. The high-frequency details of hairs and skins can help us to evaluate the fine-grained texture synthesis of models. We use 28,000 images for training and 2,000 for testing following a common setting \cite{yu2018generative,nazeri2019edgeconnect}.
	
	\item \textbf{QMUL-OpenLogo} \cite{su2018open} contains 27,083 images from 352 logo classes. Each image is annotated by fine-grained bounding box annotations of logos. We use 15,975 training images for training and 2,777 validation images for testing.
\end{itemize}

\subsection{Baselines}
\label{sec:base}

We list all the state-of-the-art models used for comparisons with their abbreviations and brief introductions as below:  
\begin{itemize}
	\item \textbf{CA} \cite{yu2018generative} is a two-stage coarse-to-fine model. It deploys a patch-based non-local module, \ie, contextual attention module, to effectively model long-range correlations.
	
	\item \textbf{PEN-Net} \cite{zeng2019learning} is built upon a U-Net \cite{ronneberger2015u} backbone. It proposes to fill missing regions from deep to shallow by  a cross-layer non-local module.
	
	\item \textbf{PConv} \cite{liu2018image} adopts a proposed partial convolution layer instead of the standard convolution to deal with the issue of color discrepancy inside and outside holes. 
	
	\item \textbf{EdgeConnect} \cite{nazeri2019edgeconnect} model consists of two stages. The first stage completes the edges of corrupted images, and the second stage completes the color images with the guidance of the completed edges from the first stage.
	
	\item \textbf{GatedConv} \cite{yu2019free} is a two-stage model that incorporates a gated convolution and SN-PatchGAN for image inpainting.  
	
	\item \textbf{HiFill} \cite{yi2020contextual} proposes a light-weight model with an efficient Contextual Residual Aggregation mechanism that enables ultra super-resolution image inpainting. 
	
	\item \textbf{MNPS} \cite{yang2017high} is a multi-scale neural patch synthesis approach based on the joint optimization of image content and texture constraints. It shows sharper and more coherent results than previous works for high-resolution images. 
\end{itemize}


\begin{table*}[!t]
	\caption{Quantitative comparisons of the proposed \modelname~with state-of-the-art deep inpainting models, \ie, CA \cite{yu2018generative}, PEN-Net \cite{zeng2019learning}, PConv \cite{liu2018image}, EdgeConnect \cite{nazeri2019edgeconnect}, GatedConv \cite{yu2019free} and HiFill \cite{yi2020contextual} on Places2 \cite{zhou2018places}. $\downarrow$ Lower is better. $\uparrow$ Higher is better. Best and second best results are \textbf{highlighted} and \underline{underlined}.}
  \label{tb:exp}
  \centering
	\resizebox{0.95\textwidth}{!}{
		\begin{tabular}{r|r|c|c|c|c|c|c|c}  \hline
		\multicolumn{2}{r|}{Mask}  
		&CA \cite{yu2018generative} &PEN-Net \cite{zeng2019learning}  &PConv \cite{liu2018image} &EdgeConnect \cite{nazeri2019edgeconnect} &GatedConv \cite{yu2019free} &HiFill \cite{yi2020contextual}  &AOT-GAN (ours) \\\hline\hline
		\multirow{6}{*}{\rotatebox{90}{$L_1(10^{-2})\downarrow$}}  
		
		&1-10\%    &0.89 &0.69 &0.68 &\underline{0.55} &0.66 &0.67 &\textbf{0.55}  \\\cline{2-9}
		&10-20\%   &2.07 &1.58 &\underline{1.28} &\underline{1.28} &1.55 &1.81 &\textbf{1.19}  \\\cline{2-9}
		~&20-30\%  &3.54 &2.72 &\underline{2.42} &\underline{2.42} &2.73 &3.31 &\textbf{2.11}  \\\cline{2-9}
		~&30-40\%  &5.19 &3.97 &\underline{3.43} &3.47 &4.08 &5.02 &\textbf{3.20}  \\\cline{2-9}
		~&40-50\%  &7.07 &5.40 &\underline{4.62} &4.92 &5.68 &7.12 &\textbf{4.51}  \\\cline{2-9}
		~&50-60\%  &10.11 &7.76 &\underline{7.74} &7.83 &8.09 &10.47 &\textbf{7.07}  \\\hline\hline
		
		\multirow{6}{*}{\rotatebox{90}{PSNR$\uparrow$}}  
		&1-10\%    &31.07 &32.73 &\underline{34.04} &33.98 &32.95 &30.97 &\textbf{34.79} \\\cline{2-9}
		&10-20\%   &25.81 &27.17 &\underline{28.75} &28.53 &27.05 &25.36 &\textbf{29.49} \\\cline{2-9}
		~&20-30\%  &22.93 &24.18 &\underline{25.59} &25.22 &23.81 &22.35 &\textbf{26.03} \\\cline{2-9}
		~&30-40\%  &20.89 &22.19 &\underline{23.40} &22.88 &21.55 &20.21 &\textbf{23.58} \\\cline{2-9}
		~&40-50\%  &19.23 &20.60 &\underline{21.56} &21.00 &19.75 &18.35 &\textbf{21.65} \\\cline{2-9}
		~&50-60\%  &17.10 &18.62 &\underline{18.75} &18.22 &16.94 &16.07 &\textbf{19.01} \\\hline\hline
		
		\multirow{6}{*}{\rotatebox{90}{SSIM$\uparrow$}}  
		&1-10\%    &0.961 &0.967 &0.971 &\underline{0.974} &0.970 &0.963 &\textbf{0.976} \\\cline{2-9}
		&10-20\%   &0.906 &0.917 &0.928 &\underline{0.931} &0.921 &0.905  &\textbf{0.940} \\\cline{2-9}
		~&20-30\%  &0.844 &0.856 &0.875 &\underline{0.876} &0.860 &0.835  &\textbf{0.890} \\\cline{2-9}
		~&30-40\%  &0.783 &0.795 &\underline{0.820} &0.816 &0.796 &0.762  &\textbf{0.835} \\\cline{2-9}
		~&40-50\%  &0.720 &0.733 &\underline{0.762} &0.751 &0.727 &0.680  &\textbf{0.773} \\\cline{2-9}
		~&50-60\%  &0.648 &0.664 &\underline{0.677} &0.660 &0.626 &0.588  &\textbf{0.682} \\\hline\hline
		
		\multirow{6}{*}{\rotatebox{90}{FID$\downarrow$}}  
		&1-10\%    &1.30 &0.42 &0.36 &0.25 &\underline{0.21} &0.53 &\textbf{0.20}\\\cline{2-9}
		&10-20\%   &6.33 &2.14 &1.85 &0.97 &\underline{0.85} &2.52 &\textbf{0.61}\\\cline{2-9}
		~&20-30\%  &17.36 &6.66 &5.83 &2.83 &\underline{2.40} &7.60 &\textbf{1.57}\\\cline{2-9}
		~&30-40\%  &34.26 &14.46 &12.96 &6.52 &\underline{5.33} &17.18 &\textbf{3.38}\\\cline{2-9}
		~&40-50\%  &56.89 &27.04 &24.63 &13.44 &\underline{10.66} &36.23 &\textbf{6.89} \\\cline{2-9}
		~&50-60\%  &82.67 &45.27 &47.09 &33.82 &\underline{32.90} &72.03 &\textbf{20.20} \\\hline
	\end{tabular}
	}
\end{table*}

\begin{figure*}[!t]
	\centering
	\includegraphics[width=\linewidth]{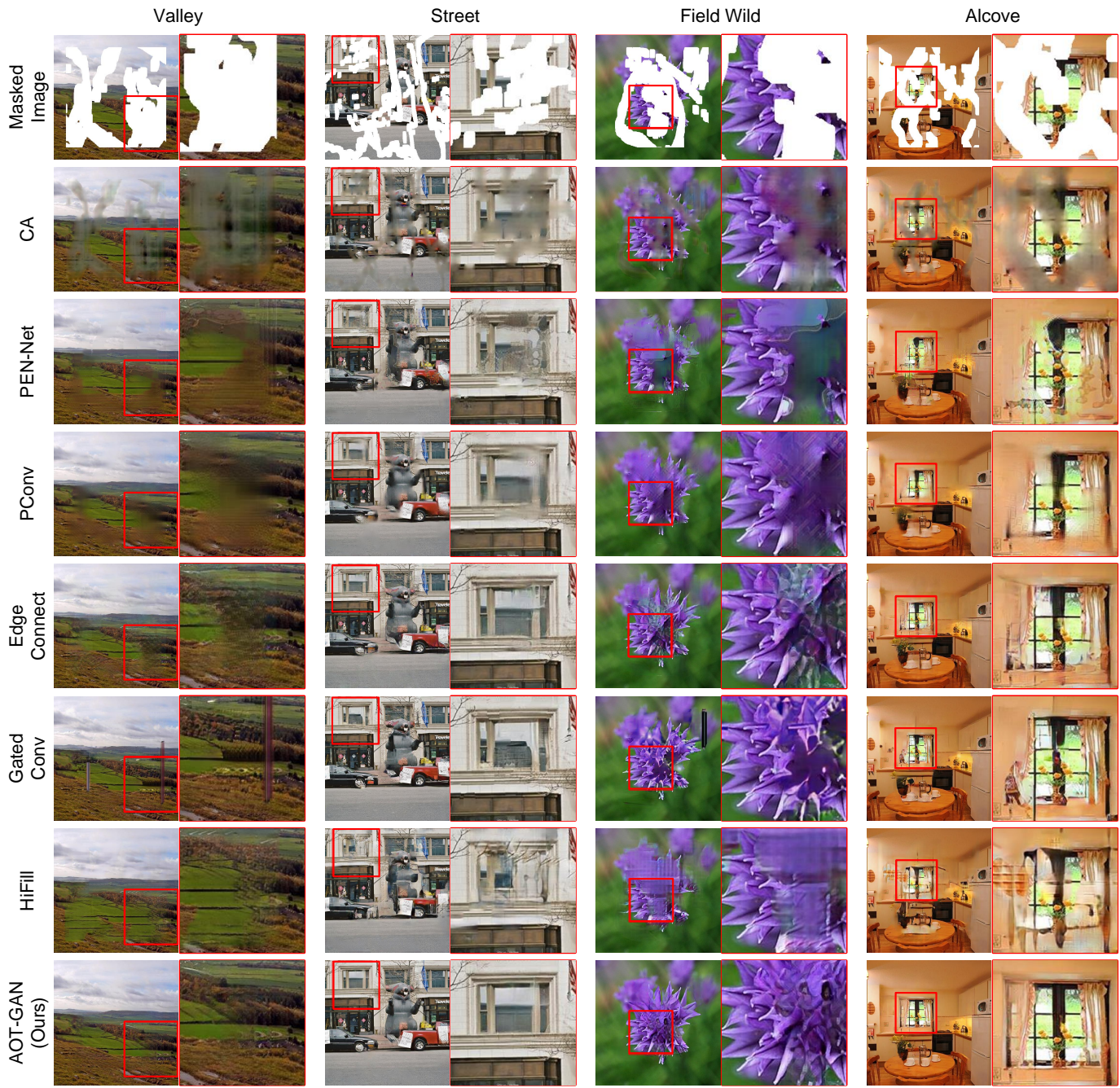}
	  \caption{Qualitative comparisons of \modelname~with CA \cite{yu2018generative}, PEN-Net \cite{zeng2019learning}, PConv \cite{liu2018image}, EdgeConnect \cite{nazeri2019edgeconnect}, GatedConv \cite{yu2019free} and HiFill \cite{yi2020contextual} on Places2 \cite{zhou2018places}. Each column shows the results and their enlarged patches marked by red boxes next to them. As shown in these cases, our model is able to reconstruct more plausible structures and clearer textures of various scenes, including \textit{valley}, \textit{street}, \textit{field wild} and \textit{alcove}. All the images are center-cropped and resized to $512\times 512$. See analysis in Section \ref{sec:qual}. [Best viewed with zooming-in]}
	  \label{fig:places}
  \end{figure*}


\subsection{Evaluation Metrics} 
\label{sec:metric}
We list all the objective metrics used for quantitative comparisons and the reasons for using them as below: 
\begin{itemize}
    \item \textbf{$L_1$ error} is widely used in previous works \cite{yu2018generative,yang2017high,zeng2019learning}. It calculates the mean absolute error between inpainted and original images to evaluate per-pixel reconstruction accuracy. 
    
    \item \textbf{Peak Signal-to-Noise Ratio (PSNR)} is one of the classical image quality assessments that used by many inpainting approaches \cite{yu2019free,yu2018generative}. 
    
    \item \textbf{Structural Similarity Index (SSIM) \cite{wang2003multiscale}} compares inpainted results with the original images from the aspects of luminance, contrast and structure. 
    
    \item \textbf{Fr\'{e}chet Inception Distance (FID) \cite{heusel2017gans}} is a popular deep metric for perceptual rationality \cite{isola2017image,miyato2018spectral}. FID measures the distance between the distributions of real and fake image features. It is worth noting that, deep metrics have been proved to be closer to human perception \cite{zhang2018unreasonable}.
\end{itemize}
Although the above four objective metrics can reflect the performance of inpainting models partly, a subjective assessment is always the best way to evaluate inpainting models. Therefore, we conduct extensive qualitative evaluations and a user study for a comprehensive comparison.

\begin{table}[!t]
	\caption{Quantitative comparison results with MNPS \cite{yang2017high} on ImageNet \cite{deng2009imagenet}. $\downarrow$ Lower is better. $\uparrow$ Higher is better.}
	\label{tb:mnps}
	\centering
	  \resizebox{0.95\linewidth}{!}{  
		  \begin{tabular}{r|c|c|c|c} \hline
				~   &$L_1(10^{-2})\downarrow$ &PSNR$\uparrow$  &SSIM$\uparrow$  &FID$\downarrow$ \\\hline \hline 
		  MNPS\cite{yang2017high}  &3.37          &21.36     &0.821     &43.87\\\hline 
		  Ours &\textbf{3.25}  &\textbf{22.10}  &\textbf{0.840}  &\textbf{41.35} \\\hline 
	  \end{tabular}
	  }
  \end{table}

\subsection{Quantitative Comparisons}
\label{sec:quant}
To evaluate the \modelname~on benchmarks, we conduct a quantitative comparison on the most challenging benchmark, Places2 \cite{zhou2018places}. Specifically, we use all the images of the evaluation set of Places2 following the common setting used in many deep inpainting approaches \cite{nazeri2019edgeconnect,liu2018image,ren2019structureflow}. Each kind of scene in Places2 contains 100 evaluation images. For each testing image, we randomly sample a free-form mask as the testing mask, which has a specific hole-to-image area ratio (\eg, 20-30\%). The free-form masks are provided by Liu \etal~\cite{liu2018image}. We use the same image-mask pairs for all approaches for fair comparisons. 

The results of quantitative comparison can be found in Table \ref{tb:exp}. It shows that our model outperforms the state-of-the-art methods in terms of all the metrics, especially in FID. Specifically, our model outperforms GatedConv \cite{yu2019free}, the most competitive model, in terms of FID with 38.60\% relative improvement in the most challenging setting (i.e., 50-60\% ratio of holes). We highlight the improvements by \modelname~in terms of FID, as FID has been verified to be closer to human perception \cite{zhang2018unreasonable}. We can safely draw the conclusion that, \modelname~outperforms the state-of-the-art models in both accurate reconstruction and synthesizing perceptually pleasing contents.

MNPS \cite{yang2017high} is a multi-stage model designed for high-resolution image inpainting  ($512\times 512$). Since only the parameters of the MNPS model trained on ImageNet \cite{deng2009imagenet} is available, we compare MNPS and \modelname~on ImageNet. 
For fair comparisons, the \modelname~is trained on ImageNet by center square inpainting masks, following the same training setting used by MNPS. 
Images are resized to $512\times 512$ for training. 
It takes MNPS a long time to fill a missing region in a $512\times 512$ image even on GPUs, due to the hundreds of iterations for post-processing. Therefore, we randomly sample 2,000 images from the testing set of ImageNet for testing and we use a center $224\times224$ square as inpainting mask, following a common setting \cite{yang2017high,song2018contextual}. The results in Table \ref{tb:mnps} show that our single-stage model is able to gain improvements in terms of all the four metrics compared with MNPS.

\subsection{Qualitative Comparisons}
\label{sec:qual}

In addition to the quantitative evaluation, we conduct qualitative comparisons with the state-of-the-art deep inpainting models, \ie, CA \cite{yu2018generative}, PEN-Net \cite{zeng2019learning}, PConv \cite{liu2018image}, EdgeConnect \cite{nazeri2019edgeconnect}, GatedConv \cite{yu2019free} and HiFill \cite{yi2020contextual} on Places2 \cite{zhou2018places}. Specifically, we show the results of various scenes, including \textit{valley}, \textit{street}, \textit{field wild} and \textit{alcove}, with their enlarged patches in Fig. \ref{fig:places}. 
The results show that, most baselines may fail in completing complex scenes with extremely large missing regions. Specifically, CA \cite{yu2018generative}
, PEN-Net \cite{zeng2019learning} and PConv \cite{liu2018image} tend to generate blurry textures. EdgeConnect \cite{nazeri2019edgeconnect} fills in the holes with checkerboard artifacts. GatedConv \cite{yu2019free} and HiFill \cite{yi2020contextual} distort the structures in results by unreasonable  repeating patterns. 
With the enhanced generator and discriminator networks, our model is able to reconstruct more plausible contextual structures and generate clearer textures in various kinds of complex scenes. 
For example, in the results of \textit{street} in Fig. \ref{fig:places}, \modelname~is able to reconstruct the structure of the window. In the results of \textit{valley}, \modelname~is able to hallucinate plausible structures and synthesize fine-grained textures of valley even with the extremely large missing regions.

\subsection{User Study}
\label{sec:user}
We can find in the quantitative comparison in Table \ref{tb:exp} that, PConv \cite{liu2018image} wins second place in terms of $L_1$ error, PSNR and SSIM. Besides, GatedConv \cite{yu2019free} wins second place in terms of FID. Therefore, we conduct a user study by pair-wise comparisons on real data, our model and the results of these top two best baselines (i.e., PConv \cite{liu2018image} and GatedConv \cite{yu2019free}). 

In each trial, two images are shown to the volunteers in an anonymous way. Specifically, one inpainted image is sampled from our model and the other is from other corresponding results (\ie, real data, or PConv \cite{liu2018image}, or GatedConv \cite{yu2019free}). 
We randomly sample 400 cases for the experiments on Places2 and CELEBA-HQ, respectively. Participants are not informed of the locations of missing regions and they are asked to select images with better structures and more realistic textures. \textbf{More than 30 participants} are invited in the user study and we have collected \textbf{951 valid votes}. The results of the user study can be found in Table \ref{tb:user}. We can see that the \modelname~outperforms other methods in most cases. Compared with real data, the results of \modelname~can surprisingly win 22.47\% cases on CELEBA-HQ. 

\begin{table}[!t]
  \caption{User study for pair-wise comparisons with PConv \cite{liu2018image}, GatedConv \cite{yu2019free}, and real data on CELEBA-HQ \cite{karras2018pggan} and Places2 \cite{zhou2018places}. ``Percentage'' means the percentage of cases where our result is better than the other.}
	\label{tb:user}
  \centering
		\resizebox{0.98\linewidth}{!}{ 
		\begin{tabular}{c|c|c} \hline 
			Percentage  &CELEBA-HQ &Places2 \\\hline\hline
			\modelname~$>$ PConv \cite{liu2018image} &95.65\%  &97.67\%    \\\hline
			\modelname~$>$ GatedConv \cite{yu2019free}   &86.62\%   &95.08\%   \\\hline
			\modelname~$>$ Real  &22.47\%   &17.39\%   \\\hline 
		\end{tabular}
		}
\end{table}

\begin{table}[!t]
  \caption{Quantitative comparisons on CELEBA-HQ \cite{karras2018pggan} for using different numbers of branches and dilation rates in \blockname~blocks. $\downarrow$ Lower is better. $\uparrow$Higher is better.}
  \label{tb:ms}
  \centering
	\resizebox{0.5\textwidth}{!}{ 
		\begin{tabular}{c|c|c|c|c|c} \hline
  \#branch & rates   &$L_1(10^{-2})\downarrow$ &PSNR$\uparrow$  &SSIM$\uparrow$  &FID$\downarrow$ \\\hline \hline 
  1 &1      &3.65 &23.16 &0.807  &14.21\\\hline 
  1 &2      &3.59 &23.29 &0.810  &12.28\\\hline 
  2 &1,2    &3.58 &23.25 &0.812  &11.54 \\\hline 
  2 &2,8    &3.36 &23.83 &0.830  &10.56 \\\hline 
  3 &1,2,8  &3.36 &23.91 &0.831  &10.06 \\\hline 
  4 &1,2,4,8 &\textbf{3.28}  &\textbf{24.06}  &\textbf{0.834}  &\textbf{9.86} \\\hline 
	\end{tabular}
	}
\end{table}

\begin{table*}[!t]
\caption{Ablation studies for each component of the proposed \modelname. The full model (\modelname) that exploits aggregated contextual transformations, gated residual connections and \softdis~performs the best.  $\downarrow$ Lower is better. $\uparrow$Higher is better.}
\label{tb:ab}
\centering
\resizebox{0.95\textwidth}{!}{ 
\begin{tabular}{c|c|c|c|c|c|c|c} \hline 
Model & \blockname~block &Gated residual connections &\softdis &$L_1(10^{-2})\downarrow$ &PSNR$\uparrow$  &SSIM$\uparrow$  &FID$\downarrow$ \\\hline\hline
1&\checkmark & &  &3.28  &24.06  &0.834  &9.86   \\\hline
2&\checkmark &\checkmark & &3.06  &24.47  &0.849  &8.02 \\\hline
3 (\modelname)&\checkmark&\checkmark &\checkmark &\textbf{2.98} &\textbf{24.65} &\textbf{0.852}  &\textbf{7.37} \\\hline 
\end{tabular}
}
\end{table*}

\subsection{Ablation Studies}
\label{sec:ab}
In this section, we conduct three sets of ablation studies to verify the effectiveness of three components of \modelname, \ie, aggregated contextual transformations, the gated residual connections and the \softdis~discriminator. As shown in Table. \ref{tb:ab}, each set of ablation studies are conducted based on previous component. Specifically, we conduct both quantitative and qualitative comparisons on CELEBA-HQ with 40-50\% mask-to-image ratio inpainting masks.

\subsubsection{Aggregated Contextual Transformations} 
\label{sec:ab-ms}
To verify the effectiveness of the aggregated contextual transformations in \blockname~blocks, we use a model of the same network depth as \modelname~as a baseline. In particularly, we remove the components of gated residual connections and \softdis. The baseline uses standard residual connections in its building blocks and it is trained by a standard PatchGAN, so that it can be improved only by the aggregated contextual transformations. We conduct extensive comparisons for the design of aggregated contextual transformations by using different numbers of branches and dilation rates based on the baseline. 

The quantitative comparison in Table \ref{tb:ms} shows that, compared with a single-branch module, the enhanced models with aggregated contextual transformations gain significant improvements in terms of $L_1$ error, PSNR, SSIM, and FID. We can also observe that a larger number of branches with more diverse dilation rates is able to achieve larger improvements. 
We speculate that this is because more branches with more diverse dilation rates can largely enrich the patterns captured by \blockname~blocks, which is important to high-resolution image inpainting, especially in filling large missing regions. 

\begin{figure}[!t]
	\centering
	\includegraphics[width=\linewidth]{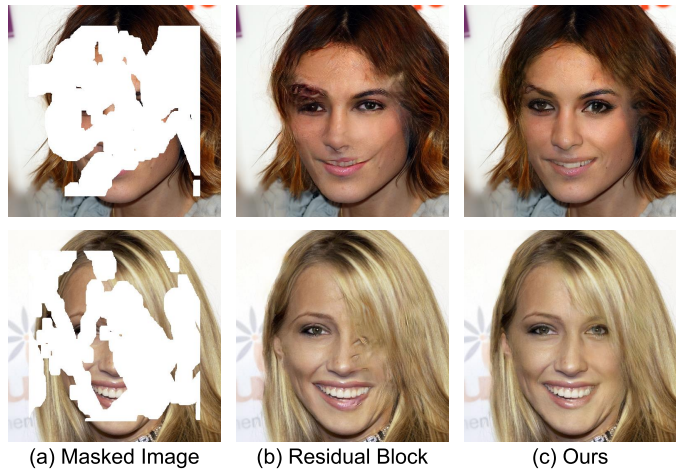}
	  \caption{Visual comparisons on images ($512\times512$) of CELEBA-HQ \cite{karras2018pggan} to verify the effectiveness of the aggregated contextual transformations in \blockname~block. [Best viewed with zoom-in]}
	  \label{fig:ms}
  \end{figure}

According to the quantitative comparisons in Table \ref{tb:ms}, we use four branches and the dilation rate of each branch is 1, 2, 4, 8, respectively in our final model. We compare this setting and the single-branch module with $rate = 2$ (\ie, the residual block used in EdgeConnect \cite{nazeri2019edgeconnect}) by qualitative comparisons in Fig. \ref{fig:ms}. We can see that, our model is able to complete the large missing region of the human face with better structures, as well as clearer textures of hairs. We can draw a conclusion that, the design of aggregated contextual transformations is able to enhance context reasoning and result in better structures and textures for high-resolution image inpainting. 

\begin{figure}[!t]
	\centering
	\includegraphics[width=\linewidth]{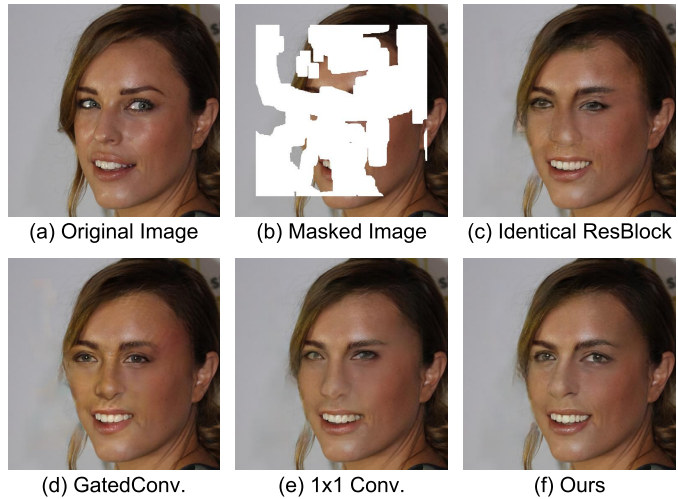}
	  \caption{Visual comparisons on images ($512\times512$) of  CELEBA-HQ \cite{karras2018pggan} to verify the effectiveness of the gated residual connection in \blockname~block. [Best viewed with zoom-in]}
	  \label{fig:gate}
  \end{figure}

  \begin{table}[!t]
	\caption{Quantitative comparisons on CELEBA-HQ \cite{karras2018pggan} to verify the effectiveness of the proposed gated residual connection in \blockname~blocks. $\downarrow$ Lower is better. $\uparrow$ Higher is better.}
	\label{tb:gate}
	\centering
	\resizebox{\linewidth}{!}
	{\begin{tabular}{r|c|c|c|c} \hline
	~  &$L_1(10^{-2})\downarrow$ &PSNR$\uparrow$ &SSIM$\uparrow$ &FID$\downarrow$ \\\hline \hline 
  Identical ResBlock &3.28  &24.06  &0.834  &9.86 \\\hline 
	GatedConv. \cite{yu2019free}  &3.28          &23.98    &0.835    &10.22 \\\hline
	$1\times 1$ conv. &3.14     &24.46    &0.846    &9.06 \\\hline
	Ours &\textbf{3.06}  &\textbf{24.47}  &\textbf{0.849}  &\textbf{8.02} \\\hline 
	  \end{tabular}}
  \end{table}

\begin{figure}[!t]
	\centering
	\includegraphics[width=\linewidth]{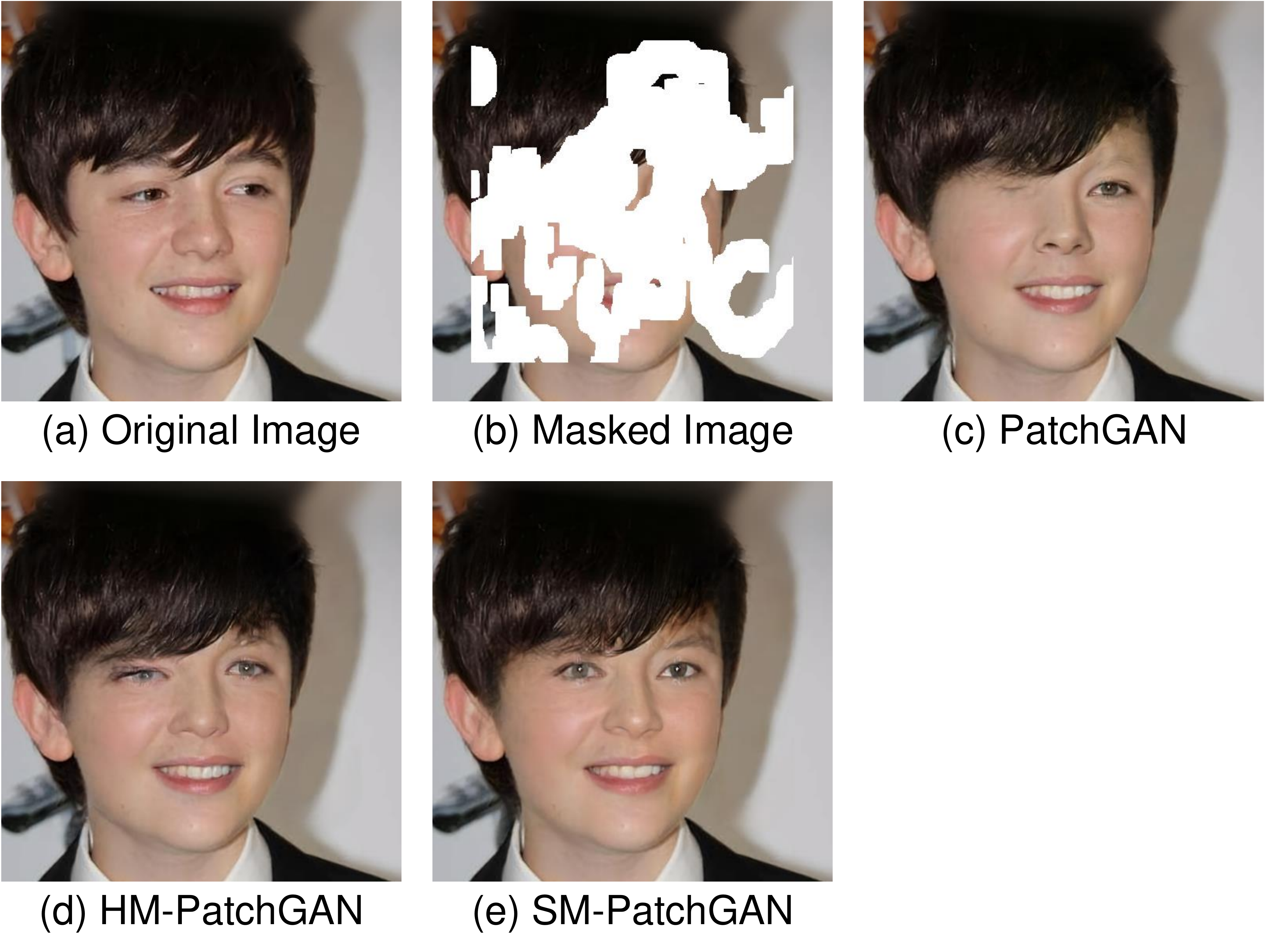}
	  \caption{Visual comparisons on images ($512\times512$) of  CELEBA-HQ \cite{karras2018pggan} to verify the effectiveness of the \softdis~discriminator of \modelname. [Best viewed with zoom-in]}
	  \label{fig:adv}
  \end{figure}

\subsubsection{Gated Residual Connections} 
\label{sec:ab-gate}
The gated residual connections are used to aggregate the input features and the learned residual features in \blockname~blocks. To verify the effectiveness of the gated residual connection, we compare it with several counterparts for aggregating residual features, \ie, identical residual connections \cite{he2016deep}, GatedConvolution \cite{yu2019free}, and feature concatenation with a $1\times 1$ convolution in \blockname~blocks. Specifically, GatedConvolution generalizes partial convolution \cite{liu2018image} by providing a learnable dynamic feature selection for each channel at each spatial location. We compare these methods by replacing the gated residual connections in \blockname~blocks. 

The quantitative comparison in Table \ref{tb:gate} shows that our gated residual connections outperforms others in terms of all the metrics. The visual comparison in Fig. \ref{fig:gate} shows that, leveraging the gated residual connections, our model is able to generate more realistic facial textures than other residual fusion methods. This is because the proposed gated residual connection provides a learnable spatially-variant connection between the input features and the learned residual features. Such a gated connection is able to update features inside missing regions, while retaining low-level details of valid pixels outside holes.

\begin{table}[!t]
  \caption{Quantitative comparisons on CELEBA-HQ \cite{karras2018pggan} to verify the effectiveness of the \softdis~discriminator of \modelname. $\downarrow$ Lower is better. $\uparrow$ Higher is better.}
  \label{tb:adv}
  \centering		
  \resizebox{\linewidth}{!}
		{\begin{tabular}{r|c|c|c|c} \hline
    ~        &$L_1(10^{-2})\downarrow$ &PSNR$\uparrow$   &SSIM$\uparrow$   &FID$\downarrow$ \\\hline \hline 
		PatchGAN \cite{isola2017image} &3.06          &24.47      &0.849      &8.02 \\\hline 
		\harddis &2.99          &24.63      &\textbf{0.853}      &7.58  \\\hline
		\softdis &\textbf{2.98} &\textbf{24.65} &0.852  &\textbf{7.37}\\\hline  
	\end{tabular}}
\end{table}

\subsubsection{Soft Mask-guided PatchGAN (\softdis)}
\label{sec:ab-dis}
We compare different discriminators in \modelname, i.e., PatchGAN \cite{isola2017image}, \harddis~and \softdis. Specifically, the generator is constructed by the proposed full \blockname~blocks. 
The quantitative comparison in Table \ref{tb:adv} shows that both the proposed \harddis~and~\softdis~outperform PathGAN. Specifically, \softdis~performs the best especially in terms of FID. 
Besides, the results of qualitative comparisons in Fig. \ref{fig:adv} show that, compared with PatchGAN and \harddis, the \softdis~is able to synthesize clearer textures of the eyes for the large missing region of the boy's face. We highlight that, the \softdis~is optimized to distinguish detailed appearances of synthesized patches of missing regions and those of contexts. Such an optimization forces the discriminator to capture realistic textures and in turn facilitates the generator to synthesize clearer textures for high-resolution image inpainting.

  \begin{figure*}[!t]
	\centering
	\includegraphics[width=\linewidth]{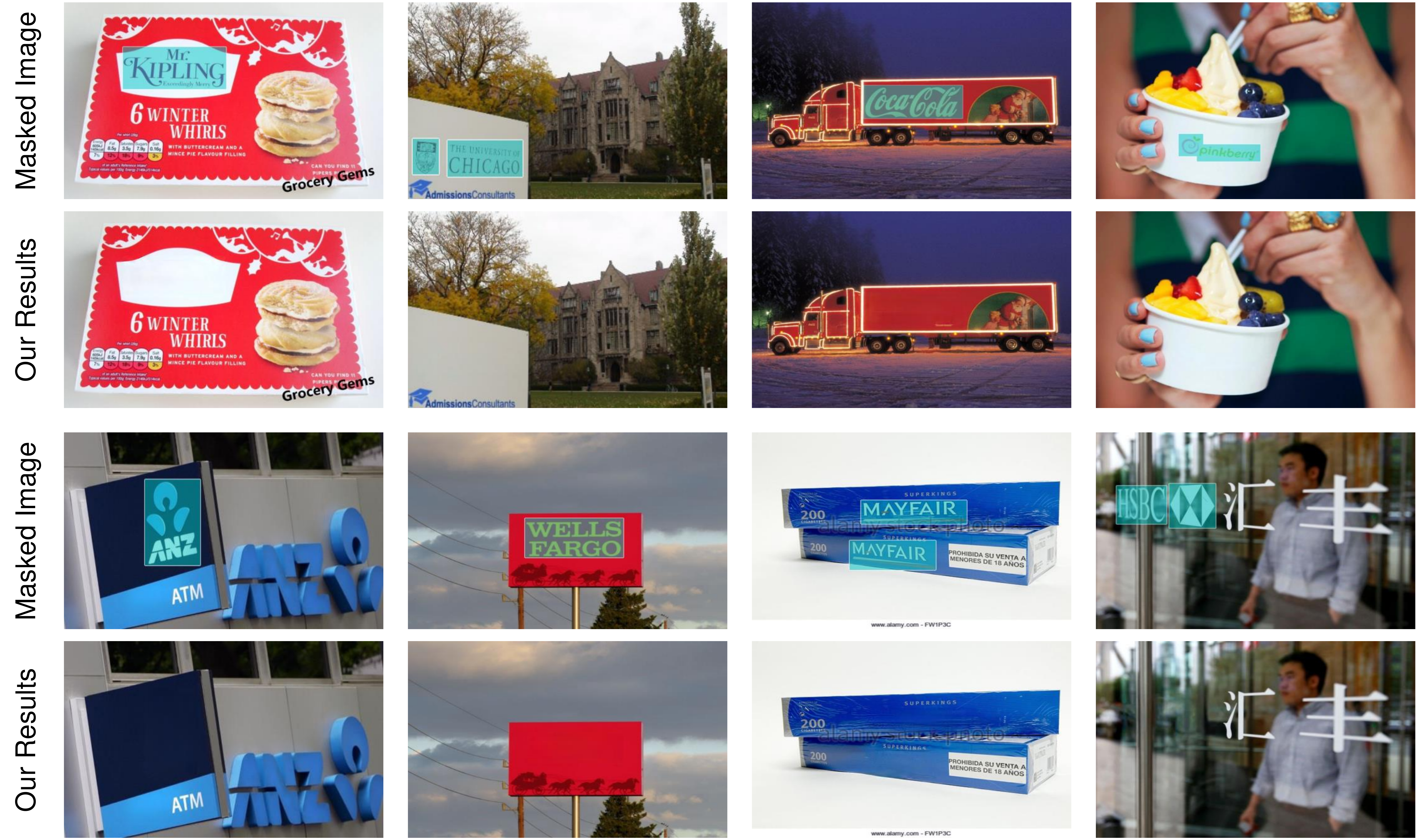}
		\caption{Visual results of the proposed \modelname~in the application of logo removal. The \modelname~is able to remove logos from images in an undetectable way and fill in missing regions with plausible contents. All these images are from QMUL-OpenLogo \cite{su2018open} and are resized to $832\times512$. We highlight the inpainting masks covering logos with blue bounding boxes. [Best viewed with zoom-in]}
		\label{fig:logo}
\end{figure*}

\subsection{Real Applications}
\label{sec:application}

\begin{figure}[!t]
\centering
\includegraphics[width=\linewidth]{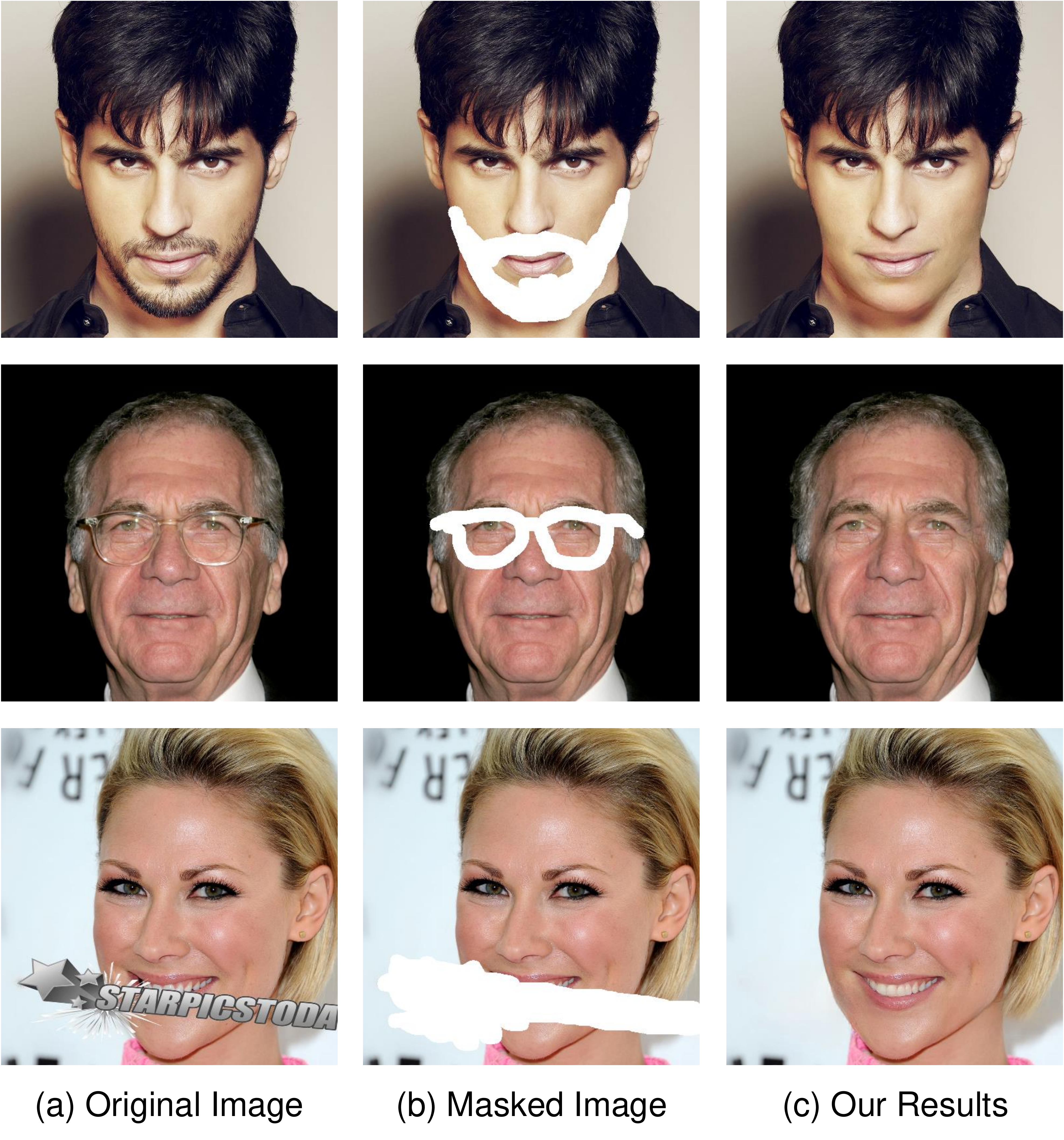}
	\caption{Visual results of \modelname~in the application of face editing. The proposed \modelname~is able to generate plausible facial structures and realistic textures for high-resolution images of faces (\ie, $512\times 512$). [Best viewed with zoom-in]}
	\label{fig:face}
\end{figure}

\begin{figure}[!t]
\centering
\includegraphics[width=\linewidth]{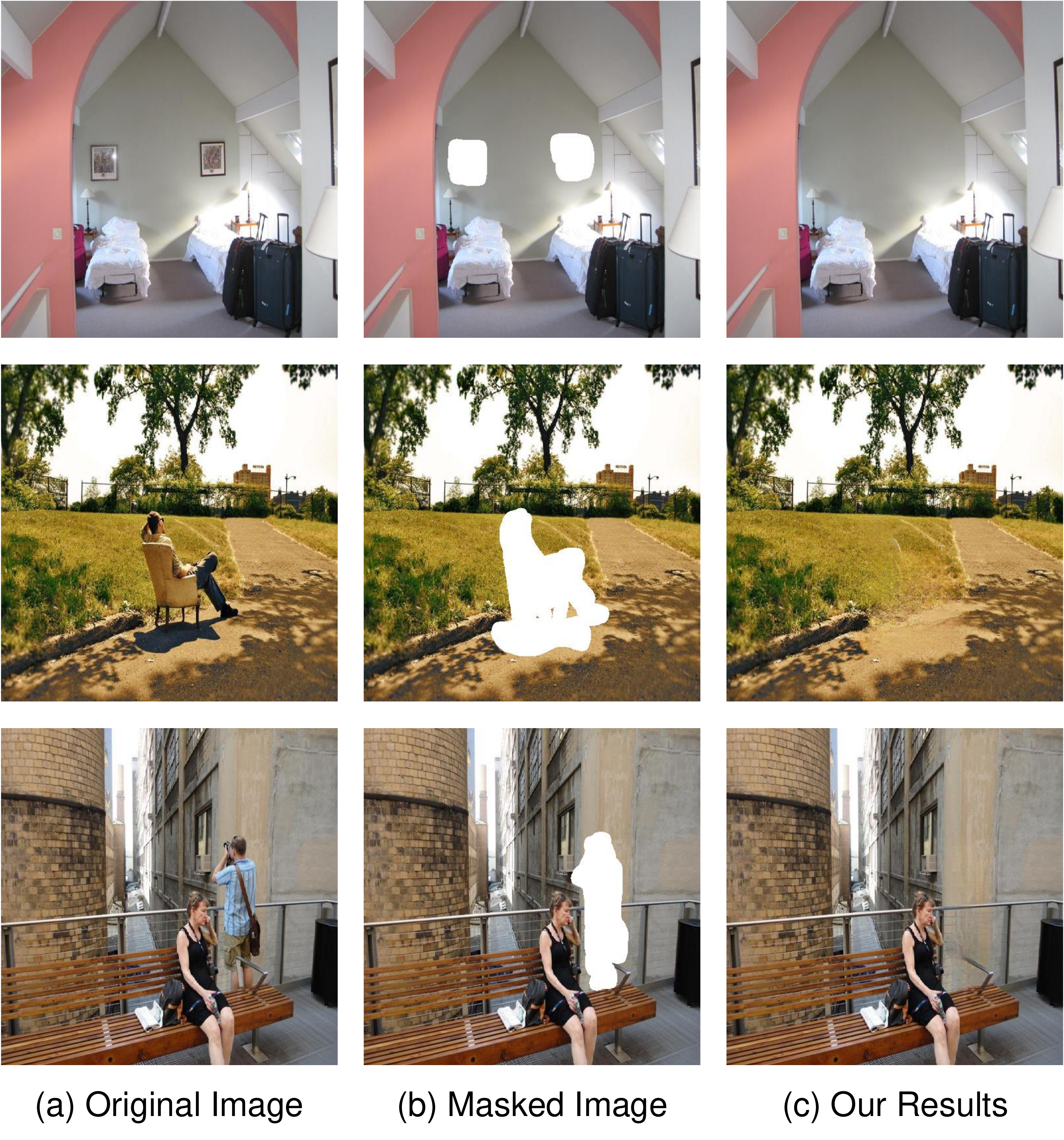}
	\caption{Visual results of the proposed \modelname~in the application of object removal. The proposed \modelname~shows promising results in filling irregular holes that provided by users in high-resolution images (\ie, $512\times 512$).  
	[Best viewed with zoom-in]}
	\label{fig:object}
\end{figure}

In this section, we apply \modelname~to three practical applications (\ie, logo removal, face editing and object removal), to verify the effectiveness of \modelname~in the real-world.

\subsubsection{Logo Removal} 
\label{sec:app-logo}
Automatic logo removal technique is very useful in logo design, media content creation, and so on \cite{yan2005automatic,qin2018visible}. It requires removing logos from images and filling in the missing regions seamlessly. However, since the scales of logos vary largely and the backgrounds are typically complex, it remains challenging for existing inpainting approaches to do well in logo removal. 
To evaluate the performance of \modelname~on logo removal, we use the images of QMUL-OpenLogo \cite{su2018open} for both training and testing.
For training, images are center-cropped and resized to $512\times512$ and we use rectangles of arbitrary sizes as training masks. For testing, images are resized to keep the shortest side as 512 and we use the bounding boxes of logos as inpainting masks. 

We show the results of \modelname~on QMUL-OpenLogo in Fig. \ref{fig:logo}. 
We can find that, \modelname~is able to remove logos from images in an undetectable way and fill in missing regions with plausible contents. 
We attribute the promising results to the enhanced context reasoning and texture synthesis by \modelname. 
Specifically, both rich patterns of interest and informative distant contexts can be captured by learning aggregated contextual transformations in the generator of \modelname, and the synthesis of clear textures are ensured by the use of \softdis.

\subsubsection{Face Editing} 
\label{sec:app-face}
The application of face editing is more and more popular in phones. Users usually use face editing to remove wrinkles, scars, or beards to make their face photo more attractive. To evaluate the performance of \modelname~on face editing, we use the images of CELEBA-HQ \cite{karras2018pggan} and all images are resized to $512\times 512$ for both training and testing. 
We show the visual results of face editing by the proposed \modelname~in Fig. \ref{fig:face}. Results show that, our model is able to complete coherent structures of faces or mouths and generate clear facial textures. For example, in the third case in Fig. \ref{fig:face}, \modelname~is able to reconstruct the missing structures of the face and the mouth of the woman face, while synthesizing fine-grained textures of the teeth.

\subsubsection{Object Removal} 
\label{sec:app-obj}

Object removal is widely used in many image editing applications. It aims at removing unwanted objects from images and fill in missing regions with plausible backgrounds. To evaluate the performance of \modelname~for object removal, we train the model on Places2 \cite{zhou2018places} with irregular masks \cite{liu2018image}. For testing, we use the masks provided by users as inpainting masks. All the images are resized to $512\times 512$ for both training and testing.
We show the results of object removal by the proposed \modelname~in Fig. \ref{fig:object}. 
The visual results show that, \modelname~is able to remove unwanted objects in different kinds of complex scenes with the real masks provided by users. For example, in the third case in Fig. \ref{fig:object}, \modelname~removes the man, reconnects the railings and synthesizes clear textures in the wall.

\subsection{Limitations}
\label{sec:limit}
We have shown that, \modelname~has achieved state-of-the-art performance with the enhanced generator and discriminator networks. In this section, we discuss two limitations of \modelname~as below. 

\textit{Customized \blockname~block.}
In practice, the number of branches and dilation rates in \blockname~block are empirically studied and set. When the image size changes, the optimal setting may need to search again. We plan to investigate adaptive mechanisms to improve \blockname~block in the future \cite{dai2017deformable,zhu2019deformable}. 

\begin{figure}[!t]
\centering
\includegraphics[width=\linewidth]{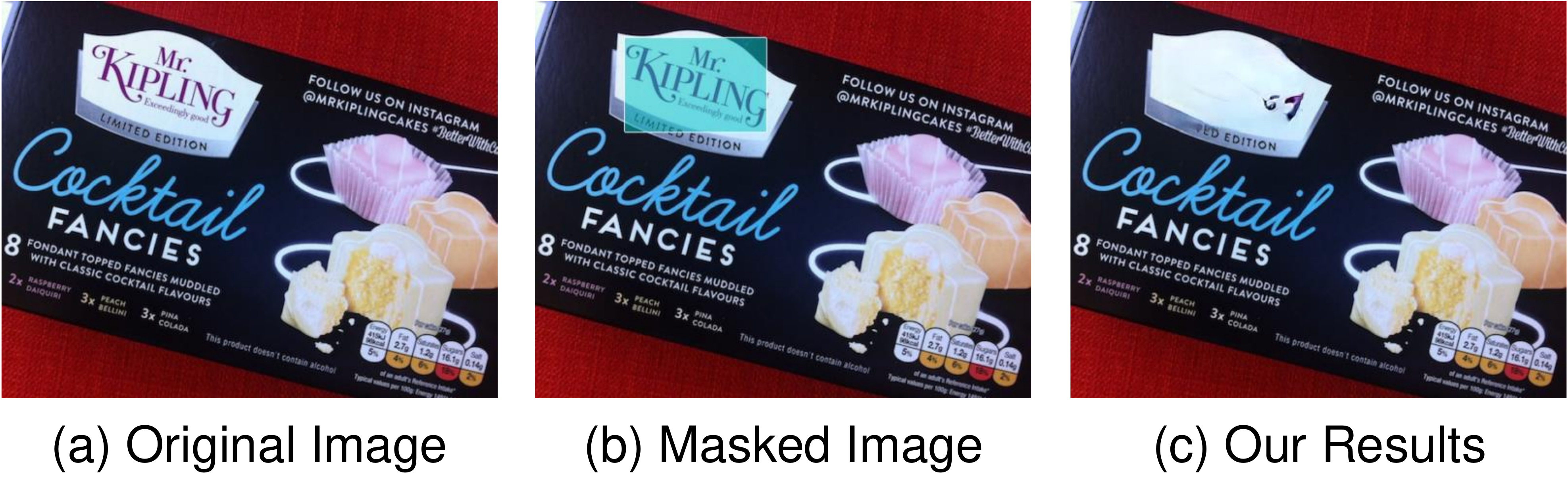}
	\caption{A failure case. When the inpainting mask can not cover the logo well and it leaves some boundaries outside the mask, \modelname~tends to propagate the boundaries and results in noticeable artifacts. [Best viewed with zoom-in]}
	\label{fig:fail}
\end{figure}

\textit{Mask selection.}
In this paper, we use user-specified foreground masks (\eg, Fig. \ref{fig:object}) or masks generated by models (\eg, Fig. \ref{fig:logo}) as our inpainting masks. However, interactive or automatic object segmentation remains a challenging problem and an active research field \cite{huang2016temporally}. As shown in the failure case in Fig. \ref{fig:fail}, when the inpainting mask can not cover the logo well and it leaves some boundaries outside the mask, it can be hard for \modelname~to remove the boundaries. Besides, it tends to propagate the boundaries and results in noticeable artifacts.  
Therefore, multiple interactive operations and refinements are required to include all affected pixels in the mask.

\section{Conclusions}
\label{sec:conclusion}
In this paper, we propose to learn aggregated contextual transformations and an enhanced discriminator for high-resolution image inpainting.
In comparison to previous approaches, the aggregated contextual transformations show impressive improvements by allowing to capture both informative distant contexts and rich patterns of interest for extremely large missing regions. Besides, the discriminator is improved by a novel mask prediction task, which in turn facilitates the generator to synthesize more realistic textures.
We conduct extensive evaluations including a user study to show that \modelname~outperforms the state-of-the-art approaches by a significant margin. Besides, we conduct ablation studies to analyze each component of \modelname. We also show the results of \modelname~on three practical applications to verify the effectiveness of \modelname~in the real-world.
As our future work, we plan to extend \modelname~to other low-level vision tasks, such as single image super-resolution, image denoising, and image-to-image translation. 

%
\section*{Acknowledgments}
This work is partially supported by NSF of China under Grant 61672548, U1611461.


\ifCLASSOPTIONcaptionsoff
  \newpage
\fi


\bibliographystyle{IEEEtran}
\bibliography{egbib}


\begin{IEEEbiography}[{\includegraphics[width=1in,height=1.25in,clip,keepaspectratio]{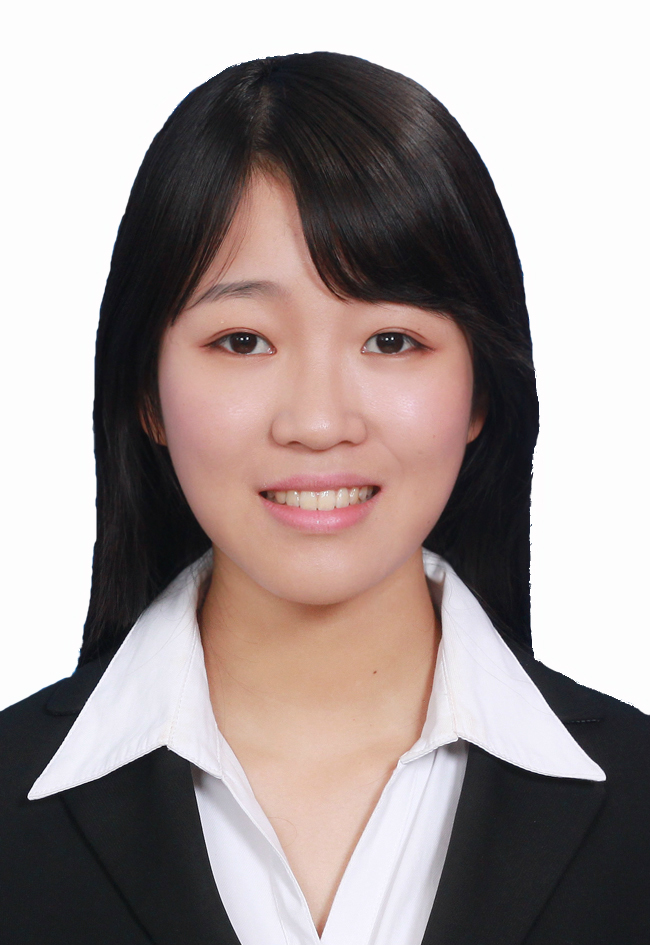}}]{Yanhong Zeng}
  Yanhong Zeng received the B.S. degree from the School of Data and Computer Science, Sun Yat-sen University, Guangzhou, China, in 2017, where she is currently pursuing the Ph.D. degree. Her research interests include image processing, deep learning and computer vision.
\end{IEEEbiography}

\begin{IEEEbiography}[{\includegraphics[width=1in,height=1.25in,clip,keepaspectratio]{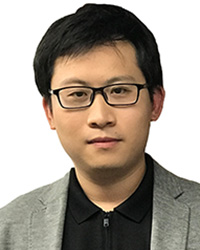}}]{Jianlong Fu}
  Jianlong Fu (M18) is currently a Lead Researcher with the Multimedia Search and Mining Group, Microsoft Research Asia (MGPA). He received his Ph.D. degree in pattern recognition and intelligent system from the Institute of Automation, Chinese Academy of Science in 2015. His current research interests include computer vision, computational photography, vision and language. He has authored or coauthored more than 40 papers in journals and conferences, and 1 book chapter. He serves as a Lead organizer and guest editor of IEEE Trans. Pattern Analysis and Machine Intelligence Special Issue on Fine-grained Categorization. He is an area chair of ACM Multimedia 2018, ICME 2019. He received the Best Paper Award from ACM Multimedia 2018, and has shipped core technologies to a number of Microsoft products, including Windows, Office, Bing Multimedia Search, Azure Media Service and XiaoIce.
\end{IEEEbiography}

\begin{IEEEbiography}[{\includegraphics[width=1in,height=1.25in,clip,keepaspectratio]{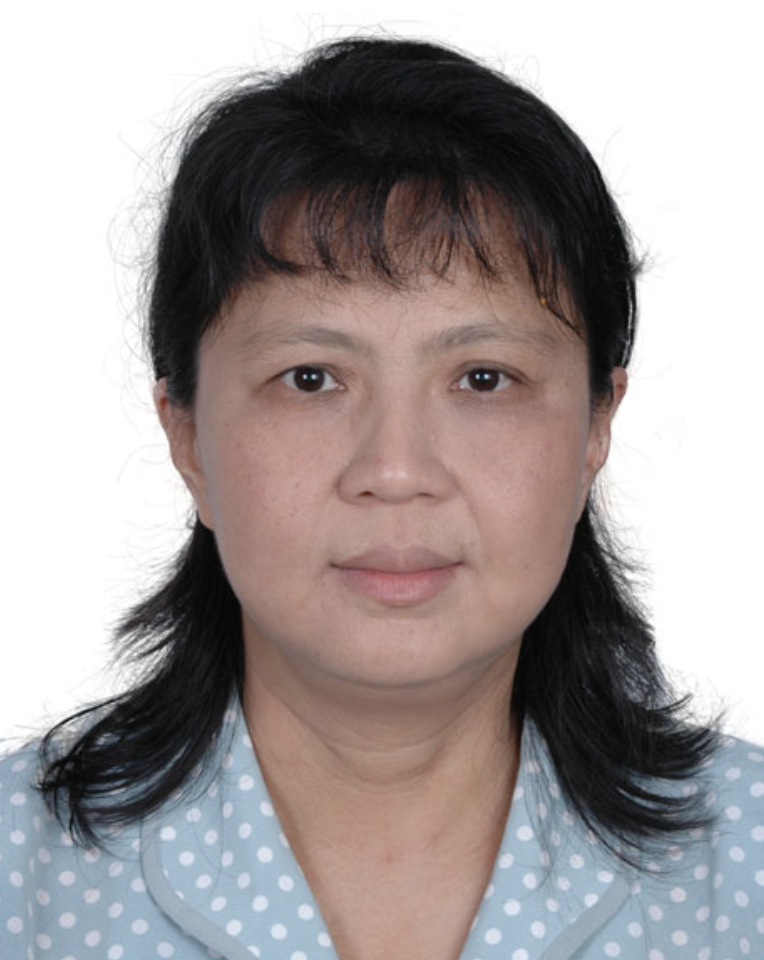}}]{Hongyang Chao}
  Hongyang Chao received the B.S. and Ph.D. degrees in computational mathematics from Sun Yet-sen University, Guangzhou, China. In 1988, she joined the Department of Computer Science, Sun Yet-sen University, where she was initially an Assistant Professor and later became an Associate Professor. She is currently a Full Professor with the School of Computer Science. She has published extensively in the area of image/video processing and holds three U.S. patents and four Chinese patents in the related area. Her current research interests include image and video processing, image and video compression, massive multimedia data analysis, and content-based image (video) retrieval.
\end{IEEEbiography}

\begin{IEEEbiography}[{\includegraphics[width=1in,height=1.25in,clip,keepaspectratio]{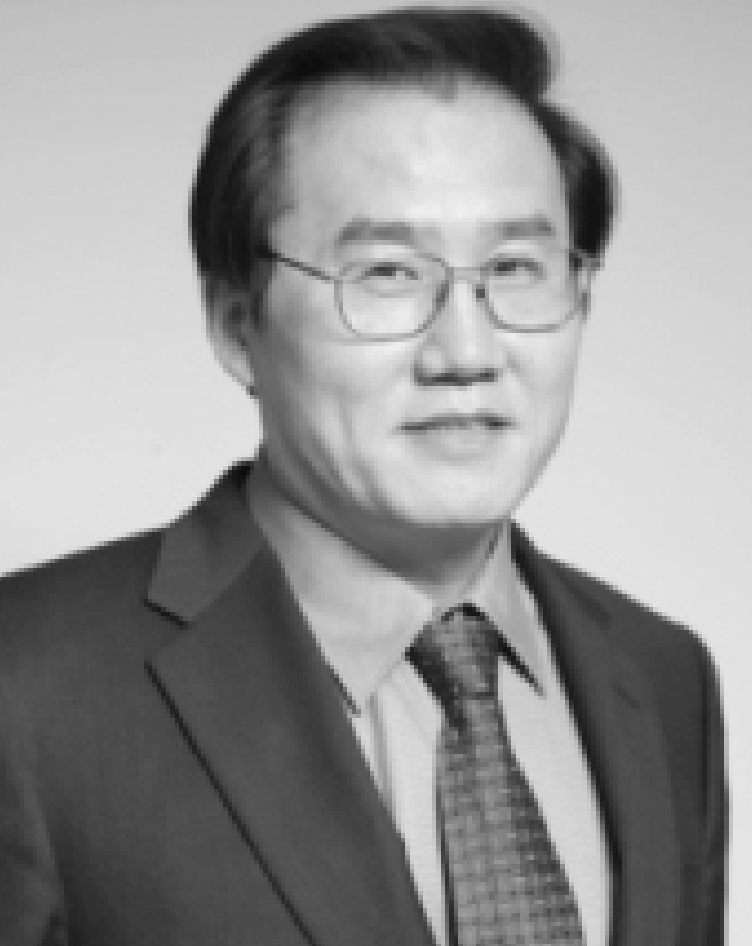}}]{Baining Guo}
  Baining Guo received the BS degree from Beijing University, and the MS and PhD Degrees from Cornell University. He is the assistant managing director of Microsoft Research Asia, where he also serves as the head of the graphics lab. Prior to joining Microsoft in 1999, he was a senior staff researcher with the Microcomputer Research Labs of Intel Corporation in Santa Clara, California. His research interests include computer graphics and visualization, in the areas of texture and reflectance modeling, texture mapping, translucent surface appearance, real-time rendering, and geometry modeling.
\end{IEEEbiography}
  
\end{document}